\documentclass{article} 
\usepackage{nips13submit_e,times}
\usepackage{hyperref}
\usepackage{url}
\usepackage{amssymb, amsmath}
\usepackage{epsfig}
\usepackage{array}
\usepackage{ifthen}
\usepackage{color}
\usepackage{fancyhdr}
\usepackage{graphicx}
\usepackage{multirow}

\title{Quantifying error in estimates of human brain fiber directions using Earth Mover's Distance }

\author{
Charles Y.~Zheng \\
Department of Statistics\\
Stanford University\\
Stanford, CA 94305 \\
\texttt{snarles@stanford.edu} \\
\And
Franco ~Pestilli \\
Department of Psychological and Brain Sciences\\
Indiana University\\
Bloomington, IN 47405\\
\texttt{franpest@indiana.edu} \\
\AND
Ariel ~Rokem \\
Department of Psychology\\
Stanford University\\
Stanford, CA 94305 \\
\texttt{arokem@stanford.edu} \\
}

\nipsfinalcopy 

\begin{document}

\maketitle

\begin{abstract} 
 Diffusion-weighted MR imaging (DWI) is the only method we
  currently have to measure connections between different parts of the
  human brain \emph{in vivo}.  To elucidate the structure of these
  connections, algorithms for tracking bundles of axonal fibers
  through the subcortical white matter rely on local estimates of the
  \emph{fiber orientation distribution function} (fODF) in different
  parts of the brain. These functions describe the relative abundance
  of populations of axonal fibers crossing each other in each
  location. Multiple models exist for estimating fODFs.  The quality of the
  resulting estimates can be quantified by means of a suitable measure of distance on the space of fODFs.
  However, there are multiple distance metrics that can be applied
  for this purpose, including smoothed $L_p$ distances and the
  Wasserstein metrics.  Here, we give four reasons for
  the use of the Earth Mover's Distance (EMD) equipped with the
  arc-length, as a distance metric.  First, the EMD is an extension of
  the intuitive angular error metric, often used in the DWI
  literature.  Second, the EMD is equally applicable to continuous
  fODFs or fODFs containing mixtures of Dirac deltas. Third, the EMD
  does not require specifying smoothing parameters.  Finally, the EMD
  is useful in practice, as well as in simulations. This is because
  the error of an estimated fODF, as quantified by the EMD of this fODF from
  the ground truth is correlated with the replicate error: the EMD between the
  fODFs calculated on two repeated measurements.
  Though we cannot calculate the error of the estimate
  directly in experimental data measured \emph{in vivo} (in contrast
  to simulation in which ground truth is known), we can use the replicate error,
  computed using repeated measurements, as a surrogate for the error. We demonstrate the application of computing the EMD-based replicate error in
  MRI data, creating anatomical contrast that is not observed with an
  estimate of model prediction error.
 \end{abstract}

\section{Introduction}

Diffusion-weighted magnetic resonance imaging (DWI) is a biomedical imaging
technique that creates images that are sensitive to the direction and distance
of water diffusion within millimeter-scale voxels in the human brain \emph{in
  vivo}.  Repeated in several different directions, diffusion sensitization can
be used to make inferences about the microstructural properties of brain
tissue in different locations, about the trajectories of organized bundles of
axons, or fascicles, and about the connectivity structure of the brain. This is
because water molecules freely diffuse along the length of nerve cell axons,
but are restricted by cell membranes and myelin along directions orthogonal to
the axon's trajectory. This technique has therefore been used in many clinical
and basic research applications [1].

To make inferences about the directions and relative fractions of
different fascicles within each region of the brain, mixture models
are employed. The signal within each volumetric pixel (or voxel) of
approximately 2x2x2 $mm^3$ is deconvolved with a kernel function,
$f_{\kappa}$, assumed to represent the signal from every individual
fascicle [2]. A set of weights, $w_i$ provides an estimate of the
fiber orientation distribution function (fODF) in each voxel, a
representation of the direction and volume fraction of different
fascicles in each voxel. However, many algorithms are proposed to
perform this deconvolution. In choosing a model and an algorithm, the
main consideration is the \emph{accuracy} of the model with respect to
the ground truth. Accuracy is defined as the average \emph{error} of
the model fit to the ground truth; error can be assessed by comparing
model fits with a known physical structure, such as excised neural
tissue that is placed in the MRI device in a particular configuration
[3]. However, direct assessment of error, and hence model accuracy, is
not applicable in human brain \emph{in vivo}.  Hence, a useful proxy
for accuracy is the \emph{precision}, or equivalently, the
\emph{reliability} of the model--how much the model fit varies due to
noise.  Precision can be estimated by fitting the model to several
replicate datasets with independent noise; computing the difference
between the fitted models produces a quantity which we refer to as the
\emph{replicate error}.  While precision does not guarantee accuracy,
the inverse statement holds--an \emph{imprecise} model will also be
\emph{inaccurate}.

Both \emph{error} and \emph{replicate error} require the specification
of a distance function or divergence on the space of fODFs.
In turn, we can quantify \emph{model inaccuracy} as average error
and \emph{model imprecision} as average replicate error.\footnote{
Note that the definition of accuracy and precision resemble
but subtly differ from the statistical concepts of \emph{bias} and \emph{variance}.
Bias refers to the difference between the average model fit and the ground truth.
However, in non-Euclidean spaces, there may not exist an operation for averaging
multiple model fits, making the concept of bias inapplicable.
Meanwhile, variance is defined as half the average \emph{squared} distance
between two model fits, in contrast to imprecision, which is
the average distance between two model fits.}
Previous DWI studies have used numerical simulations to assess the
fits of algorithms to the fODF [2, 4, 5], and the angular error (AE),
quantified as the sum of the minimal arc distances between the true
directions and estimated directions, is commonly used as a measure of
inaccuracy in these studies. AE has an intuitive appeal, but its
application to fODFs with multiple non-zero weights is problematic,
since angular error ignores the relative weights of the directions,
and also fails to penalize fODFs with an incorrect number of
directions. However, the fODF is naturally interpreted as a probability distribution of
directions.  Thus, any distance between probability distributions
could be used to measure distances between fODFs. In the present
study, we examine three commonly used distances or divergences: total
variation (TV), Kullback-Leibler divergence (KL), and earth mover's
distance (EMD). We demonstrate that the EMD has several advantages
over other measures of discrepancy between fODFs.

\section{Methods and Theory}

\subsection{Models}\label{ss:models}
We model the diffusion signal using a \emph{sparse fascicle model}
(SFM).  Originating from work by [6] and further developed by Behrens
et al, Dell'Acqua et al and Tournier et al [2, 7, 8], these models
describe every voxel of the white matter as composed of $k$ distinct
populations of fibers, where $k$ is an integer greater or equal to 1.
The directions of the fibers are unit vectors $v_1,\hdots,v_k$, and we
do not distinguish between a vector $v$ and its mirror image $-v$,
because DWI measurements are antipodally symmetric.  The weights of
the fibers are real positive numbers $w_1,\hdots,w_k$ and add to 1,
reflecting the fractional volume occupied by the fiber population.
The signal measured in direction $x_i$ is:
\[
y_i  \sim Rician(\tilde{S}_0 \sum_{j=1}^k w_j e^{-\kappa(v_j^T x_i)},\sigma^2)
\]
where $\tilde{S}_0$ is a scaling parameter, $\kappa$ is a free
parameter which is assumed constant given fixed experimental
parameters (gradient field strength, pulse duration, etc.), and the
Rician distribution [9] is defined by $ \sqrt{(\mu + Z_1)^2 + Z_2^2} \sim
Rician(\mu,\sigma^2)$ for $Z_i \stackrel{iid}{\sim} N(0,\sigma^2)$.

Under the general framework of the SFM, one arrives at more specific models by
making particular assumptions about the number of fibers and their properties.
One might assume the assumption of a particular lower bound for the angular
separation between distinct fiber populations, a minimal threshold on the
proportion of a fiber in a voxel, or an upper limit on the number of distinct
fibers in a voxel.  Furthermore, it is necessary to specify the parameter
$\kappa$; one can estimate $\kappa$ from the data, or rely on a biophysical
model. In the simulation studies, we will treat $\kappa$ and $\sigma^2$ as
known parameters.

The SFM can be formulated as a Bayesian model, by specifying priors on
the number of fibers, the directions of the fibers, and the weights of
the fibers.  For reasons of computational tractability, we assume
$k=2$ fibers and that each fiber has a weight of 0.5, with a direction
which is independently uniformly distributed.  The posterior
distribution for this model can be easily computed, by discretizing
the projective plane.  Supposing the data is also generated by the
same priors, the Bayesian posterior allows one to obtain optimal point
estimates.  However, one could consider the Bayesian model as a useful
approximation to the truth even when the priors are incorrect.

Inference of the SFM is simplified considerably if one is willing to
model the signal as having a Gaussian distribution rather than a
Rician distribution.  Under the assumption of Gaussianity, the fODF
$\hat{f}$ is estimated through non-negative least squares (NNLS):
\[
\hat{f} = \sum_{j=0}^p \frac{\beta_j}{\sum_i \beta_i}\delta_{u_j}
\]
\begin{equation}\label{obj}
\beta = \text{argmin}_{\beta > 0} \sum_{i=1}^n\left|y_i-\sum_{j=1}^p \beta_j e^{-\kappa(u_j^T x_i)}\right|^2
\end{equation}
where $u_1,\hdots,u_p$ are points from an arbitrarily fine sampling of
the projective plane.\footnote{ It is common to apply
  regularization, such as an $L_1$ penalty [2], or elastic net penalty
  [12], to the objective function \eqref{obj}.  However, NNLS yields
  useful estimates even without regularization; hence we neglect the
  regularized variants of NNLS in this paper.}  The NNLS method does
not constrain the number of directions with positive weights.
However, one can choose to use best-$\hat{K}$-subset regression
(B$\hat{K}$S)
\footnote{Finding the best set of $\hat{K}$ directions is an NP-hard
  problem in general.  However, two considerations make it feasible in
  the application of DWI imaging.  One, there are scientific reasons
  to assume that $\hat{K}$ is a small number, e.g. from two five.
  Two, we are willing to tolerate a small angular error in the chosen
  directions.  These two factors mean that a brute force search is
  possible, though still computationally expensive.  This is in
  contrast to the general problem of best subset regression, which
  often requires the use of greedy search or convex approximation.}
to constrain the number of directions to $\hat{K}$: $\beta =
\text{argmin}_{\beta > 0, ||\beta||_0 = \hat{K}}
\sum_{i=1}^n\left|y_i-\sum_{j=1}^p \beta_j e^{-\kappa(u_j^T
  x_i)}\right|^2$. Here $||\cdot||_0$ is the $L_0$ pseudonorm, which
counts the number of nonzero components in the vector.

\begin{figure}[htbp]
\centering
\includegraphics[scale=0.4, trim = 50mm 50mm 50mm 50mm, clip]{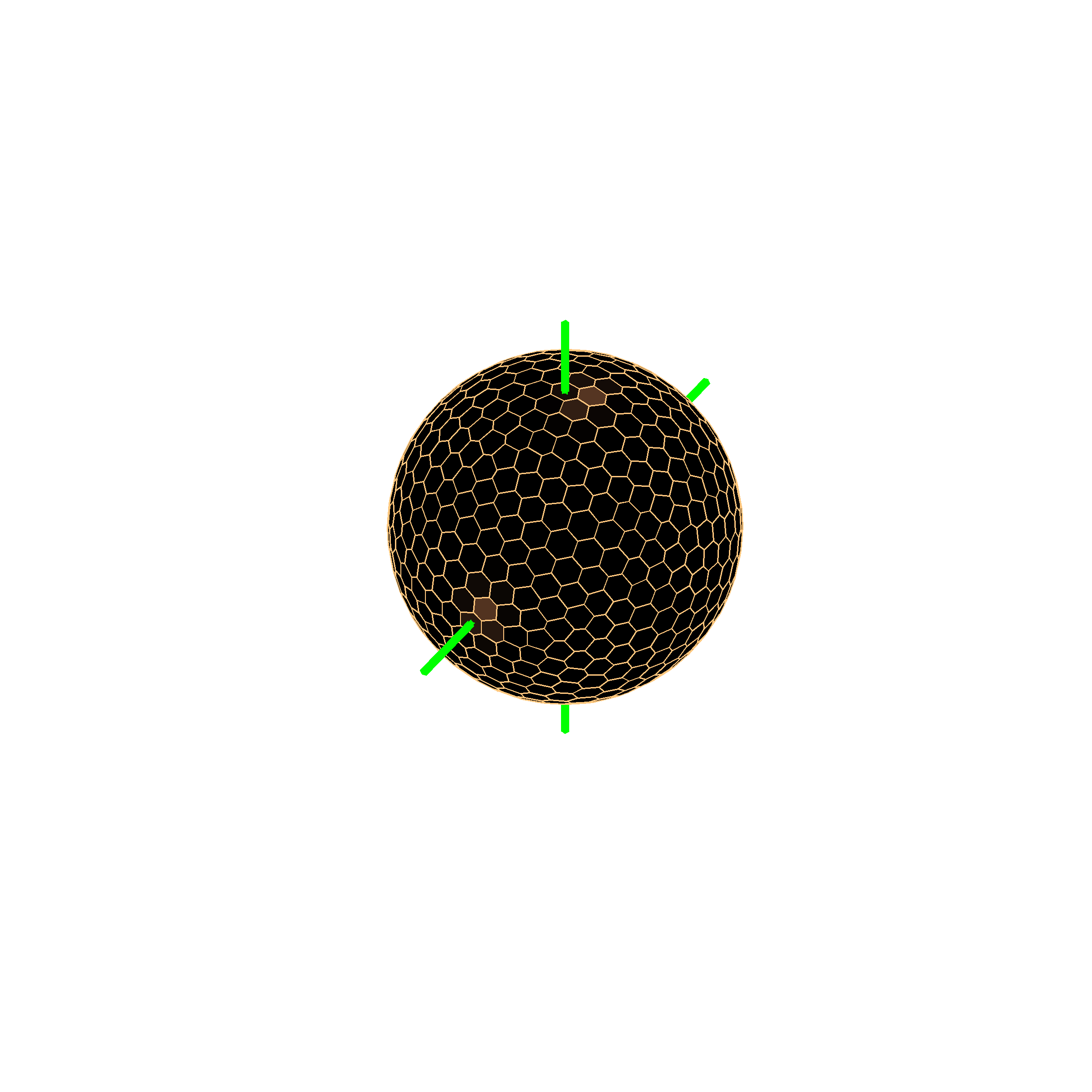}
\includegraphics[scale=0.4, trim = 50mm 50mm 50mm 50mm, clip]{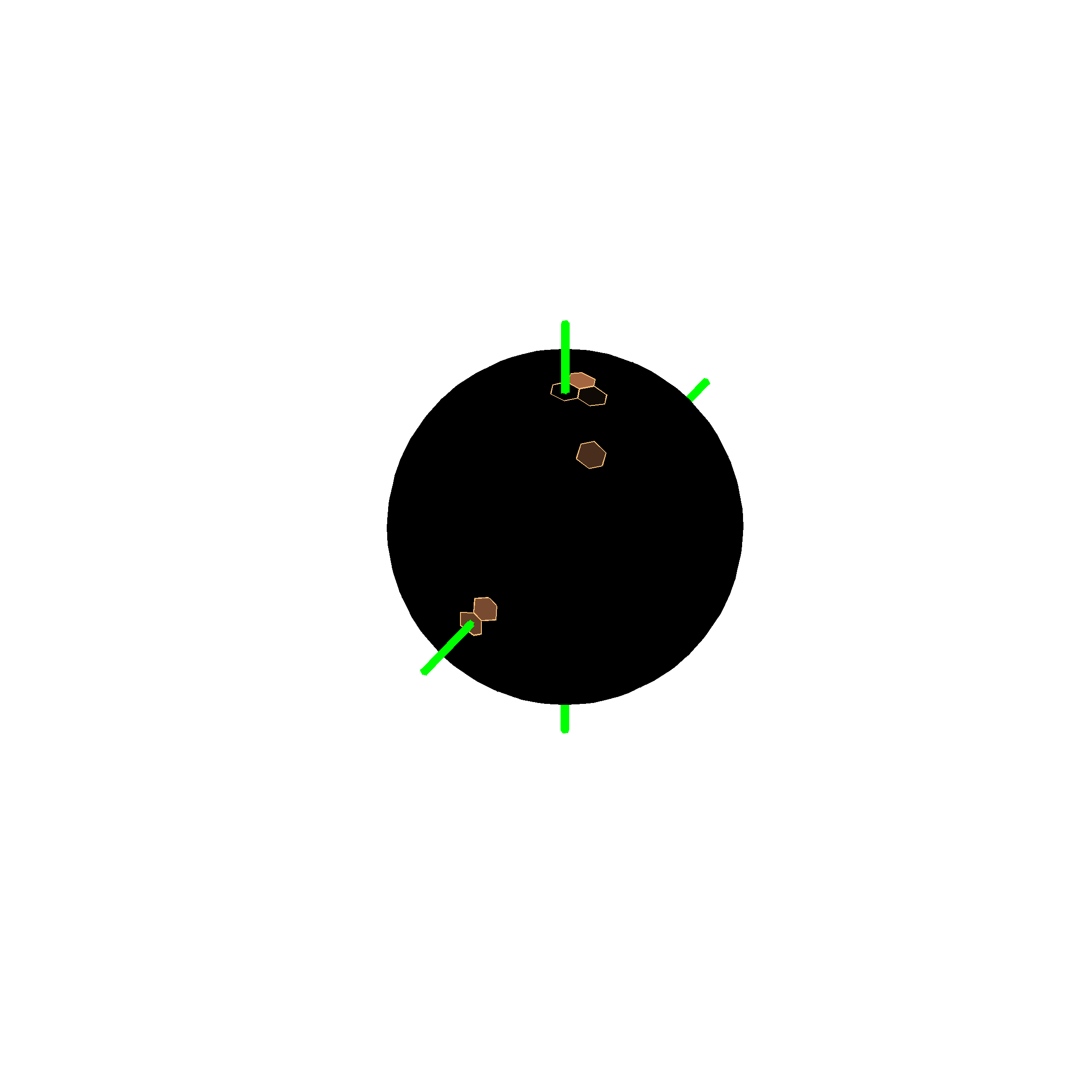}
\includegraphics[scale=0.4, trim = 50mm 50mm 50mm 50mm, clip]{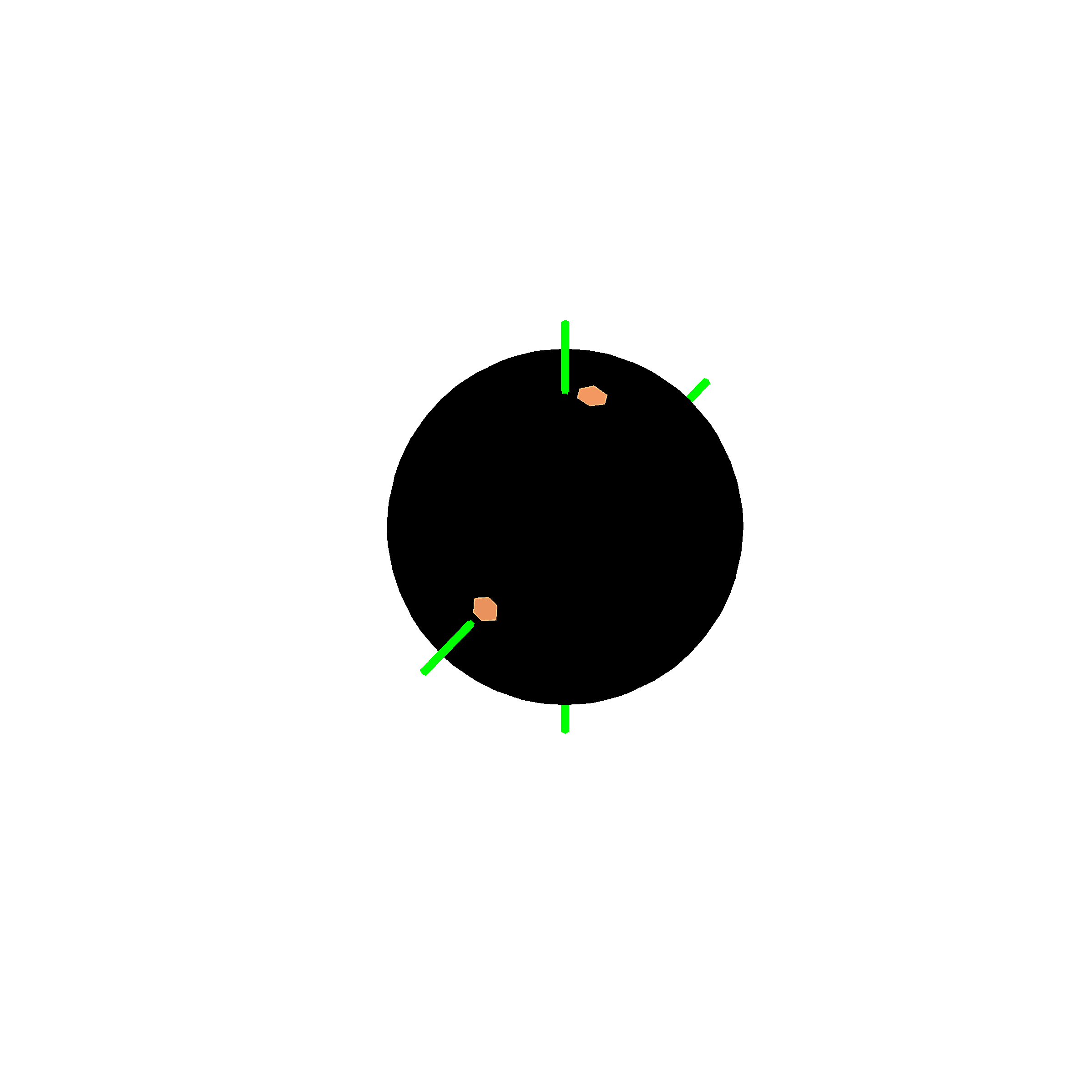}
\caption{fODFs estimated by the Bayesian posterior mean (left), NNLS (center),
  and Best-2-subset regression (B2S; right) given the same data.  Sticks
  indicate true directions entered into simulation. Light edges on polygons
  indicate positive probability mass. At the one end, the Bayesian posterior
  mean is continuous, while at the other, B2S produces a very sparse estimate,
  with NNLS having intermediate sparsity. }
\label{fig:model_demo}
\end{figure}

Figure \ref{fig:model_demo} illustrates example fODFs estimated by the Bayesian
posterior mean, NNLS and best-2-subset regression (B2S).

\subsection{Distances for Probability Distributions}\label{ss:distances}

\begin{figure}[htbp]
\centering
\includegraphics[scale=0.18]{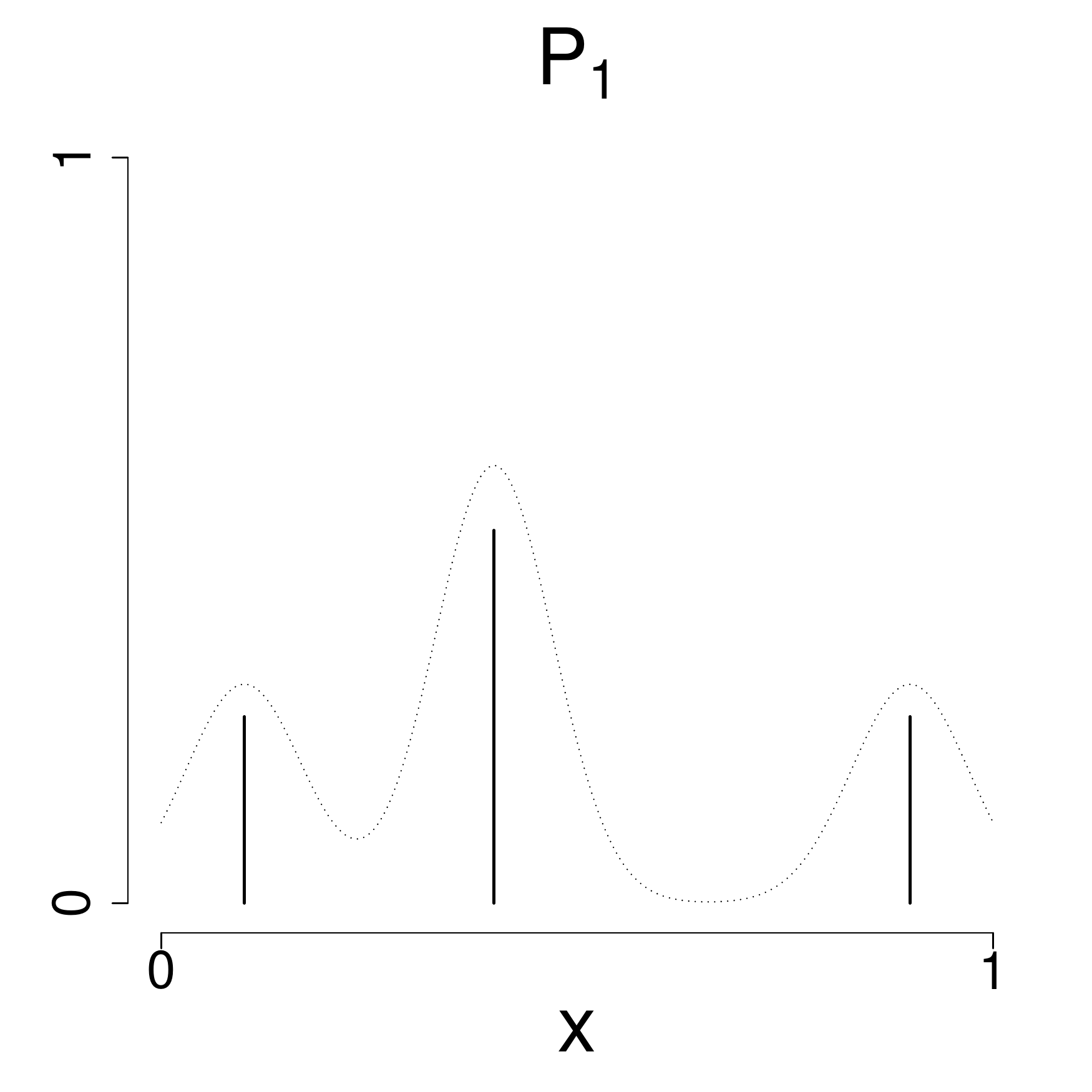}
\includegraphics[scale=0.18]{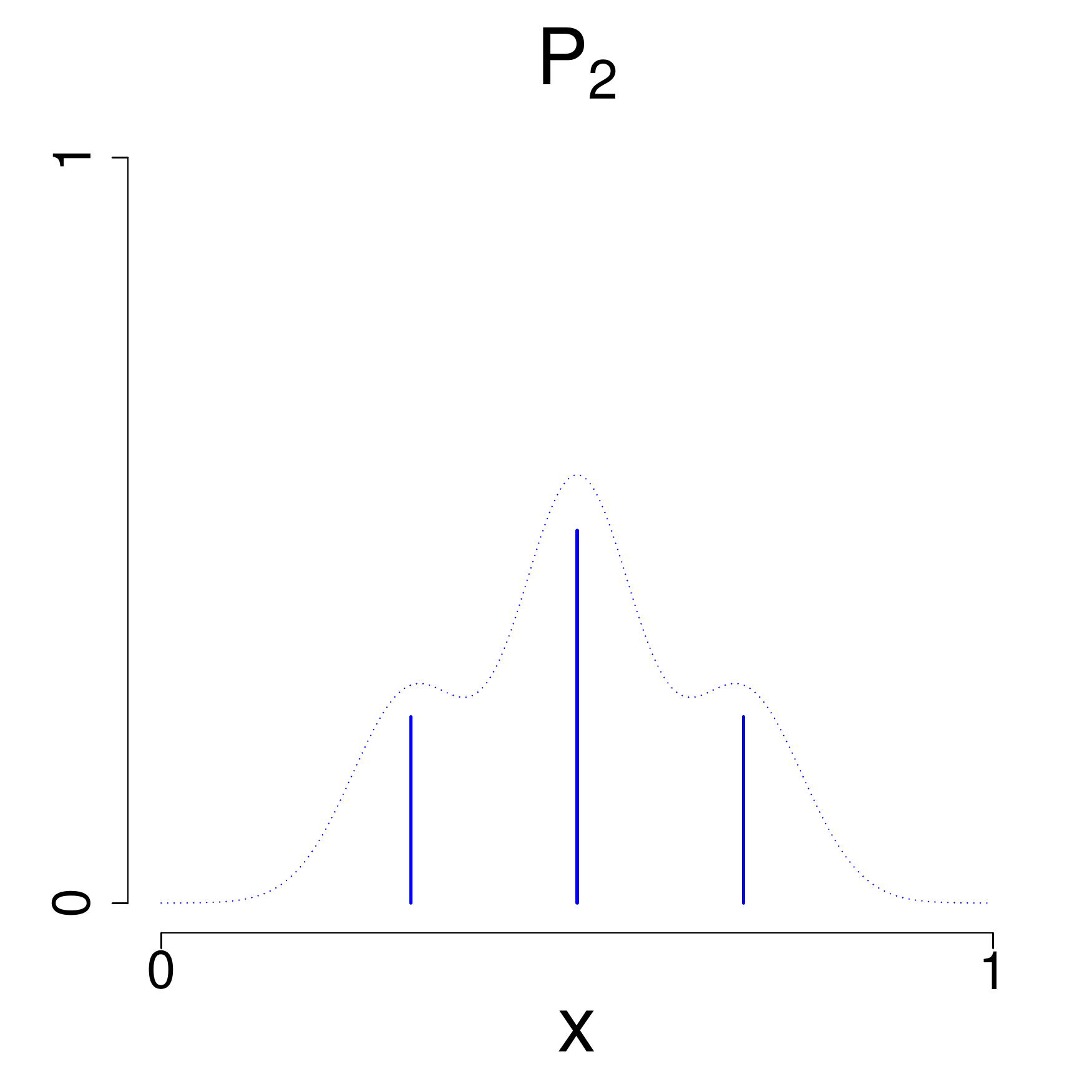}
\includegraphics[scale=0.18]{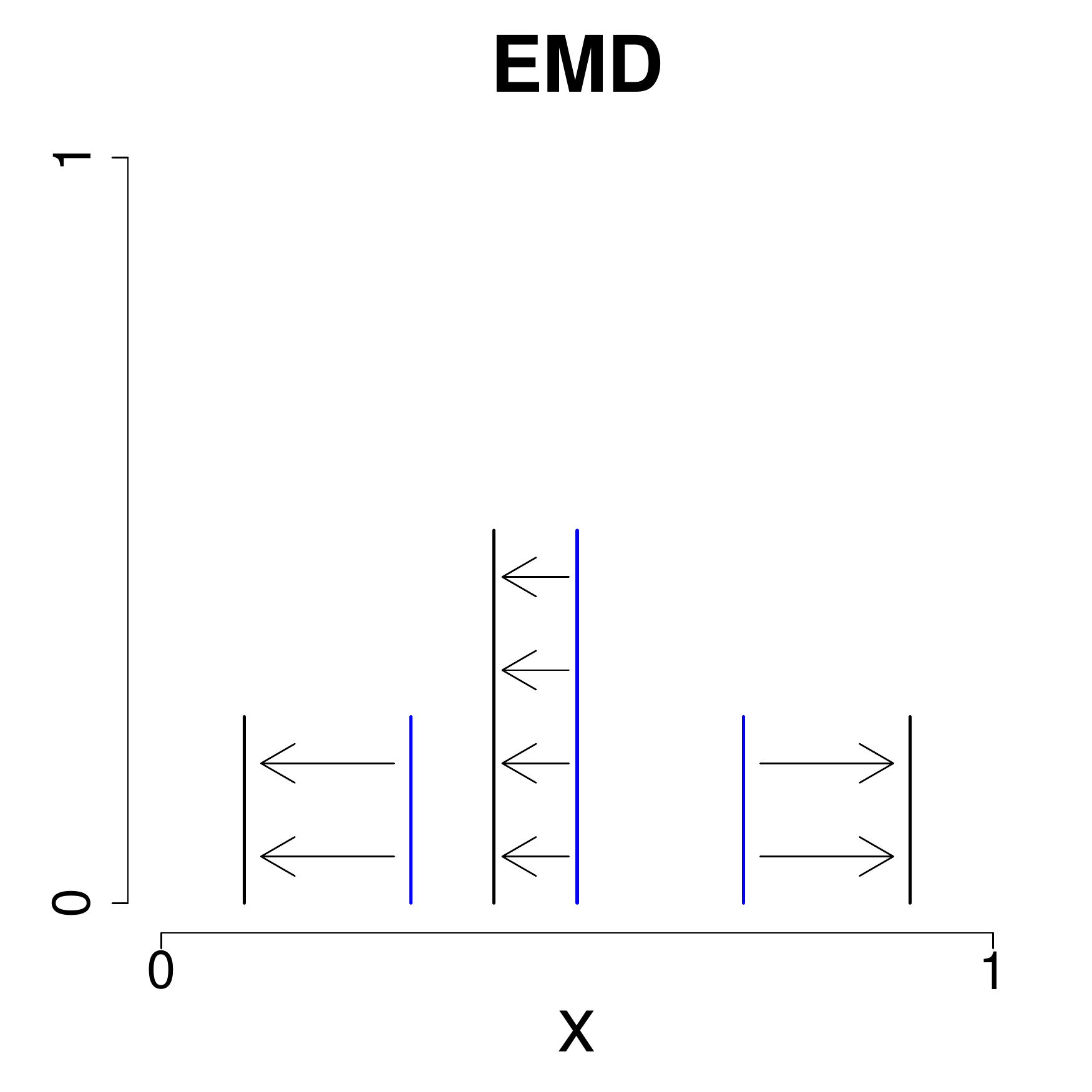}
\includegraphics[scale=0.18]{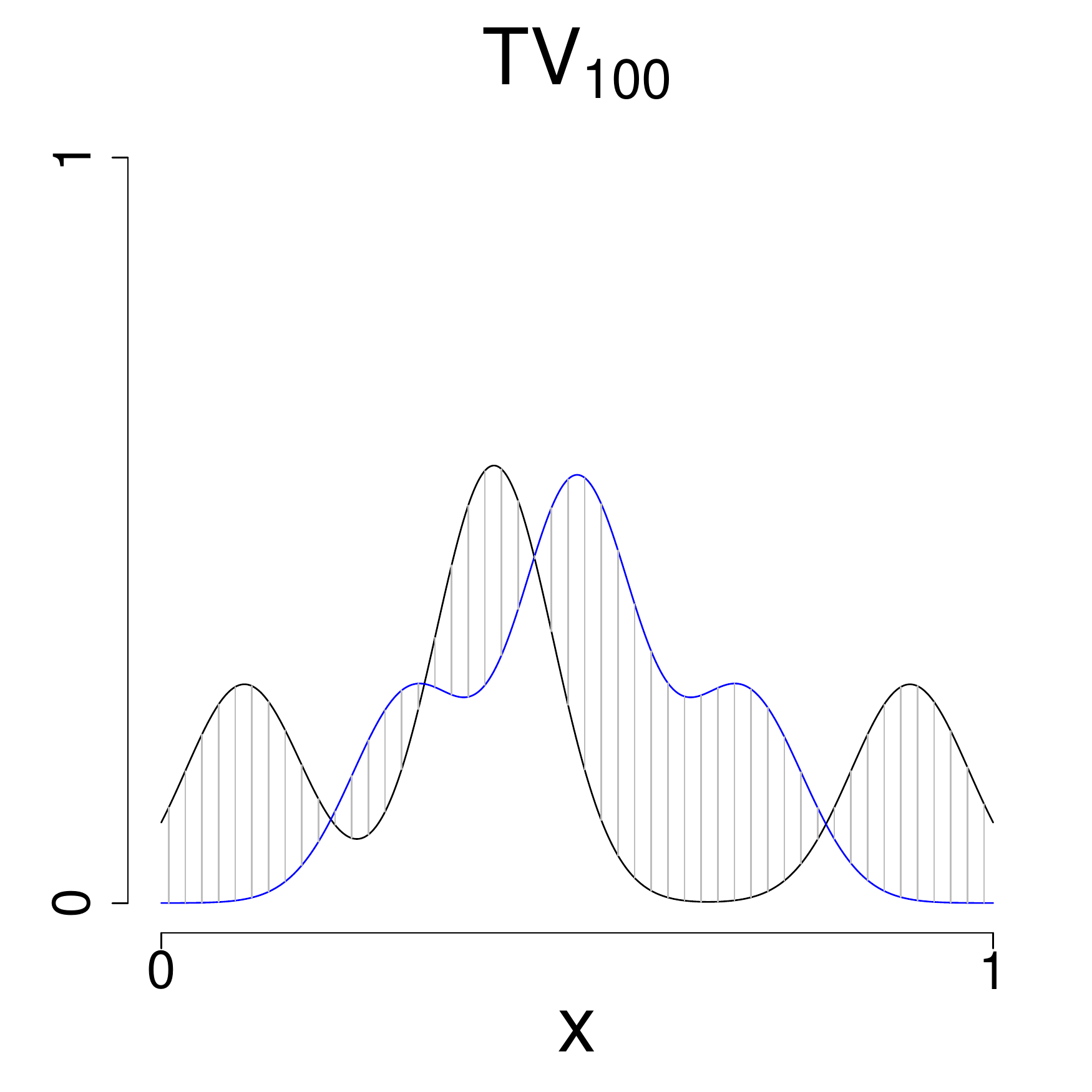}
\caption{Schematic illustrating the difference between EMD and
  kernel-smoothed TV distance.  Left to right: (i) probability
  distribution $P_1$, a mixture of Dirac deltas (solid spikes), with
  kernel-smoothed form $P_{1,gauss,100}$ superimposed (dottted lines);
  (ii) distribution $P_2$ with smoothed form $P_{2,gauss,100}$
  superimposed; (iii) computation of $d_{EMD}(P_1,P_2)$; (iv)
  computation of $d_{TV,gauss,100}(P_1,P_2) =
  d_{TV}(P_{1,gauss,100},P_{2,gauss,100})$}
\label{fig:schematic}
\end{figure}

The total variation metric for distributions $P$ and $Q$ is defined as
$d_{TV}(P,Q) = \sup_A |P(A)-Q(A)|$ where $A$ is an arbitrary
measurable set and is easy to compute: Given vectors $p$ and $q$ which
are histograms for $P$ and $Q$ respectively, TV is approximated by
$\frac{1}{2}|p-q|_1$, with $|\cdot|_1$ the $L_1$ vector norm.  Another
commonly used characterization of distance between distributions is
the Kullback-Leibler divergence: $KL(P,Q) = \int \log(dP/dQ)dP$. We
will use the symmetrized KL divergence, defined by $SKL(P,Q) =
\frac{1}{2}KL(P,Q) + \frac{1}{2}KL(Q,P)$.  Note that while TV is a
distance metric, neither KL divergence nor symmetrized KL divergence
are metric.  Both TV and KL divergence are unsuitable to compare
distributions which are mixtures of Dirac delta function.  If $P$
and $Q$ are two distributions with disjoint support, the
total variation distance $d_{TV}(P,Q)$ will be equal to 1, while
the KL divergence $KL(P,Q)$ will be infinite, regardless of how
close or how far the atoms of $P$ and $Q$ are from each
other.  Therefore, rather than applying total variation or KL
divergence directly, one can first apply \emph{kernel smoothing} to
the distributions, then compute the distance between the smoothed fODFs
[10]. Here, we use Gaussian smoothing, parameterized by $\lambda > 0$,
and we write the convolution of $P$ with the gaussian kernel with mean zero
and variance $\lambda^{-1}$ by $P \star \phi_\lambda$.
Hence we define the smoothed TV distance as $d_{TV,\lambda}(P,Q) =
d_{TV}(P \star \phi_\lambda,Q \star \phi_\lambda)$ and the smoothed
symmetrized KL divergence as $SKL_\lambda(P,Q) =
SKL(P \star \phi_\lambda,Q \star \phi_\lambda)$.  Figure
\ref{fig:schematic} illustrates the calculation of smoothed TV
distance in a one-dimensional setting.

The earth mover's distance (EMD), or 1-Wasserstein distance, can be
interpreted as the minimal amount of work needed to transform $P_1$
into $P_2$, by optimally transporting the mass from $P_1$ to the mass
in $P_2$. The work is measured by the total distance times mass
transported; a general definition can be found in [11].  Figure
\ref{fig:schematic} illustrates the calculation of EMD in a
one-dimensional setting.  In contrast to the TV distance or the KL
divergence, the EMD depends on the notion of a distance or cost
between two points: in other words, it incorporates the geometry of
the underlying space.  The EMD between two distributions $P_1$ and $P_2$ can
be computed by linear programming in the special case that $P_1$ and
$P_2$ are mixtures of Dirac deltas; i.e., $P_i = \sum_{j=1}^{k^i}
w^i_j \delta_{v^i_j}$.  Then
\begin{equation}\label{eq:emd}
d_{EMD}(P_1,P_2) = \min_x \sum_{i=1}^{k^1}\sum_{j=1}^{k^2} c_{ij}x_{ij} \text{ subject to }
x_{ij} \geq 0,\ \ 
\sum_{i=1}^{k^1} x_{ij} = w_j^2,\ \
\sum_{j=1}^{k^2} x_{ij} = w_i^1
\end{equation}
where $c_{ij} = d(v_i^1,v_j^2)$ for a suitable distance metric $d$.
Here $x_{ij}$ is understood as the amount of mass moved from the point
$v_i^1$ in $P_1$ to the point $v_i^2$ in $P_2$.

The 2-Wasserstein distance (2WD) has a similar definition to EMD,
replacing $d_{EMD}$ with $d_{Was2}^2$ and replacing $c_{ij}$ with
$c_{ij}^2$ in equation \eqref{eq:emd}.

Because the EMD and 2-Wasserstein distance (2WD) are equipped with a notion of
geometrical distance, either can be used to quantify how two mixtures of
Dirac delta functions are ``close'' even though none of the delta
functions overlap, and in contrast to the KL and TV metrics, does not
require the choice of an arbitrary smoothing parameter.  

It is possible to state a number of additional properties of the
aforementioned distance metrics when they are applied in Euclidean
space.

First is the concept of \emph{scale equivariance}.  Given a
probability distribution $P$, one can define \emph{scaling by a
  constant} $\lambda > 0$ by defining the scaling measure $\lambda P$
\[
(\lambda P)(A) = P(\frac{1}{\lambda}A)
\]
recalling that $\lambda A$ is defined as $\lambda A = \{\lambda x: x \in A\}$.
Then the property of scale equivariance is defined as
\[
d(P,Q) = \frac{1}{\lambda} d(\lambda P, \lambda Q)
\]
for all probability distributions $P$, $Q$.
It is easy to prove that EMD and 2-Wasserstein distance
satisfy scale equivariance.
Meanwhile, total variation satisfies \emph{scale invariance}
rather than scale equivariance, which is the property that
\[
d(P,Q) = d(\lambda P,\lambda Q).
\]
But smoothed total variation satisfies neither scale equivariance
nor scale invariance, due to the smoothing parameter.

The second concept is that of \emph{robustness to outliers}.
Given a probability distribution $P$ with mass concentrated in a small ball (say, the unit ball),
one can consider \emph{contamination} of the distribution with a point mass located at a distant point $x$.
That is, consider transforming $P$ to $(1-\epsilon)P + \epsilon \delta_x$ for $x$ with $||x||$ large.
The robustness of the distance metric $d$ is determined by the behavior of the quantity
\[
d(P,(1-\epsilon)P + \epsilon \delta_x)
\]
as $\epsilon \to 0$, $||x|| \to \infty$.
We have for small $\epsilon$ and large $x$ that
\begin{align*}
d_{EMD}(P,(1-\epsilon)P + \epsilon \delta_x) &\approx \epsilon ||x||\\
d_{2WD}(P,(1-\epsilon)P + \epsilon \delta_x) &\approx \sqrt{\epsilon} ||x||\\
d_{TV,\lambda}(P,(1-\epsilon)P + \epsilon \delta_x) &\approx \epsilon
\end{align*}
Both EMD and smoothed TV have a linear dependence on $\epsilon$ while 2-Wasserstein has
a square root dependence on $\epsilon$. This means that 2-Wasserstein is much less
robust to contamination for small $\epsilon$.
Meanwhile, only smoothed TV has an $O(1)$ dependence on $||x||$, meaning that
smoothed TV is the most robust to outliers.

While the two properties of scale equivariance and robustness to outliers
are only defined for Euclidean spaces, we will see that they are still useful
for understanding the properties of the distance metrics in non-Euclidean settings,
such as the projective plane.

\subsection{Distances for fODFs}

All of the aforementioned distance metrics can be adapted for the projective plane,
and thus used to measure distances between fODFs.

\begin{figure}[htbp]
\centering
\includegraphics[scale=0.30, trim = 10mm 0mm 10mm 10mm, clip]{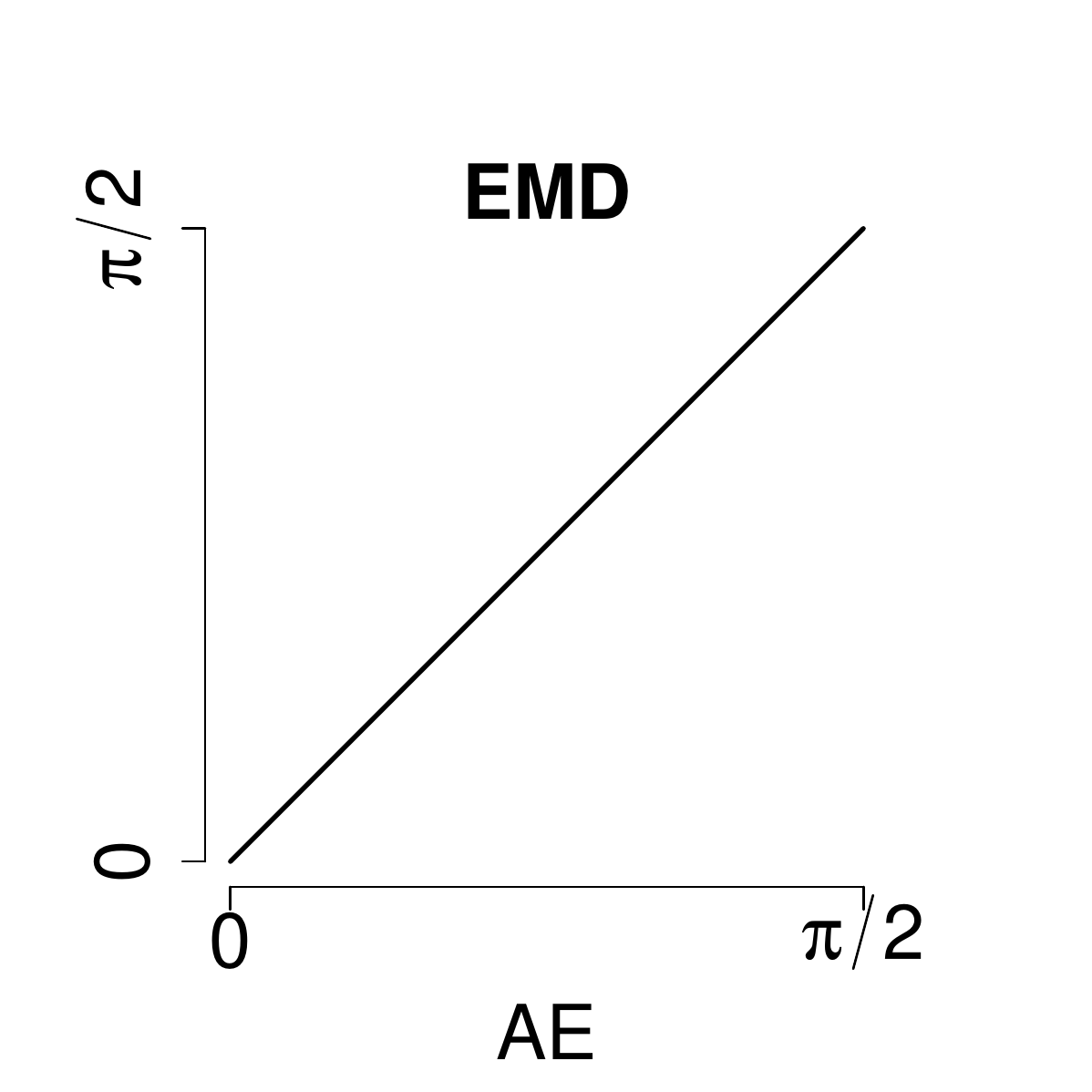}
\includegraphics[scale=0.30, trim = 10mm 0mm 10mm 10mm, clip]{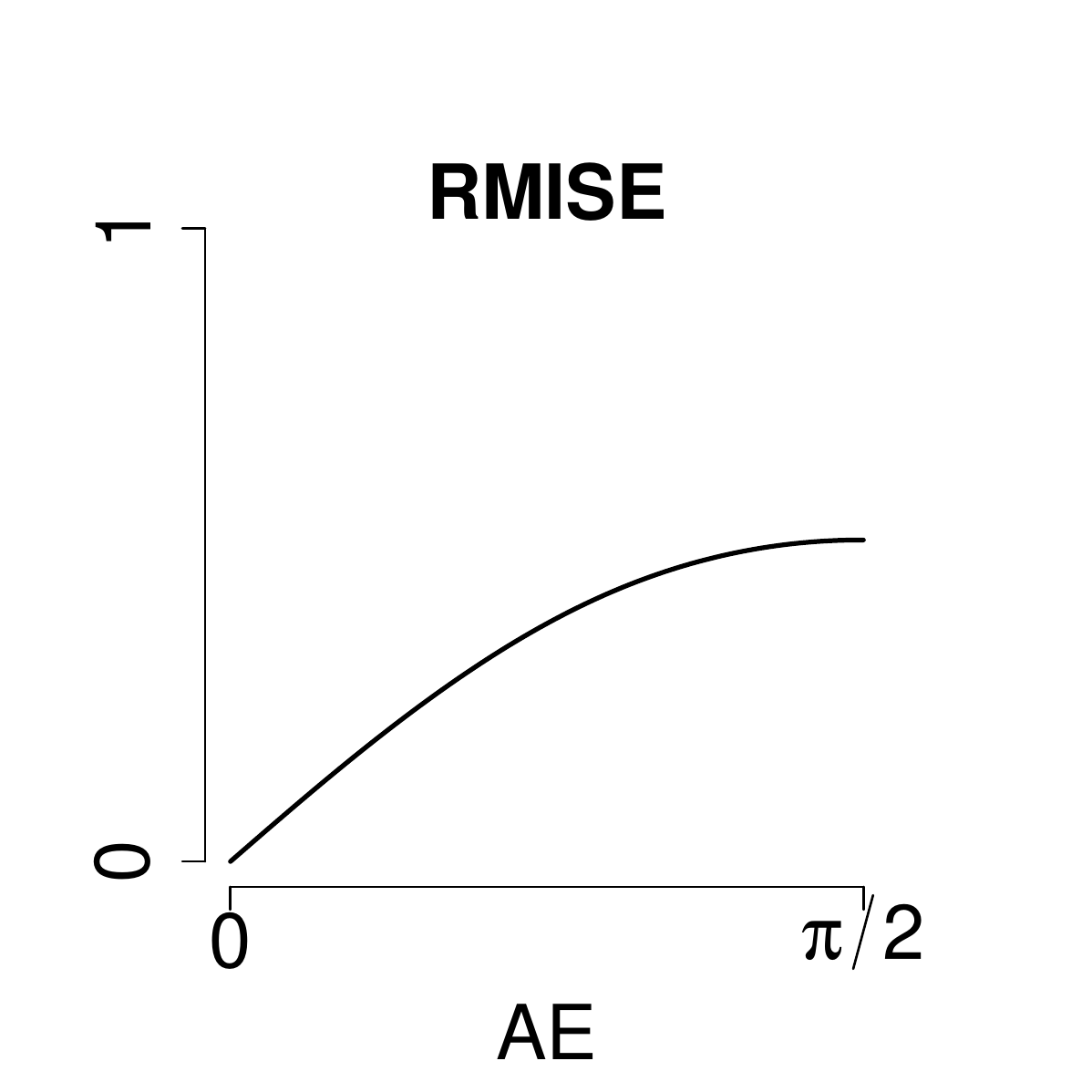}
\includegraphics[scale=0.30, trim = 10mm 0mm 10mm 10mm, clip]{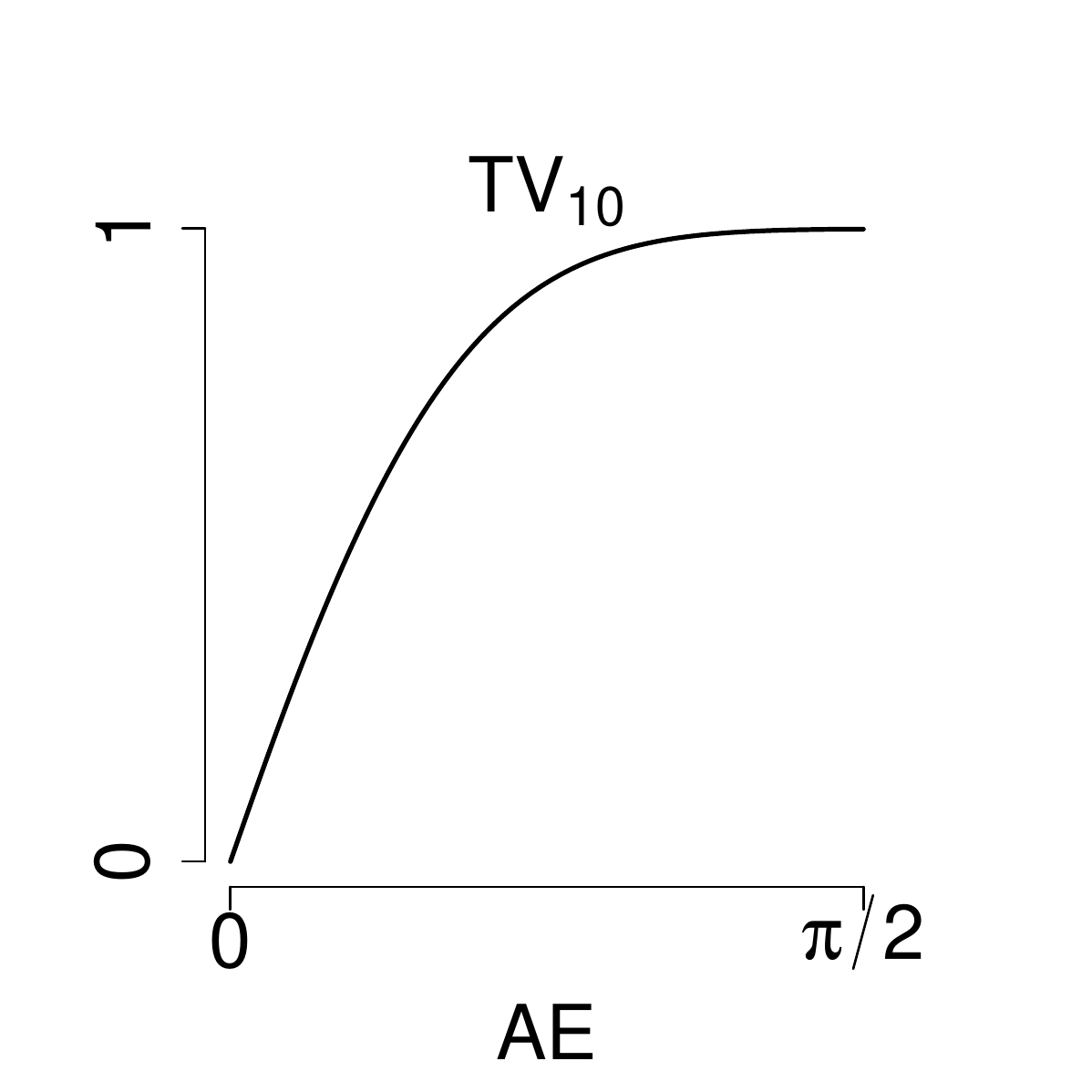}
\includegraphics[scale=0.30, trim = 10mm 0mm 10mm 10mm, clip]{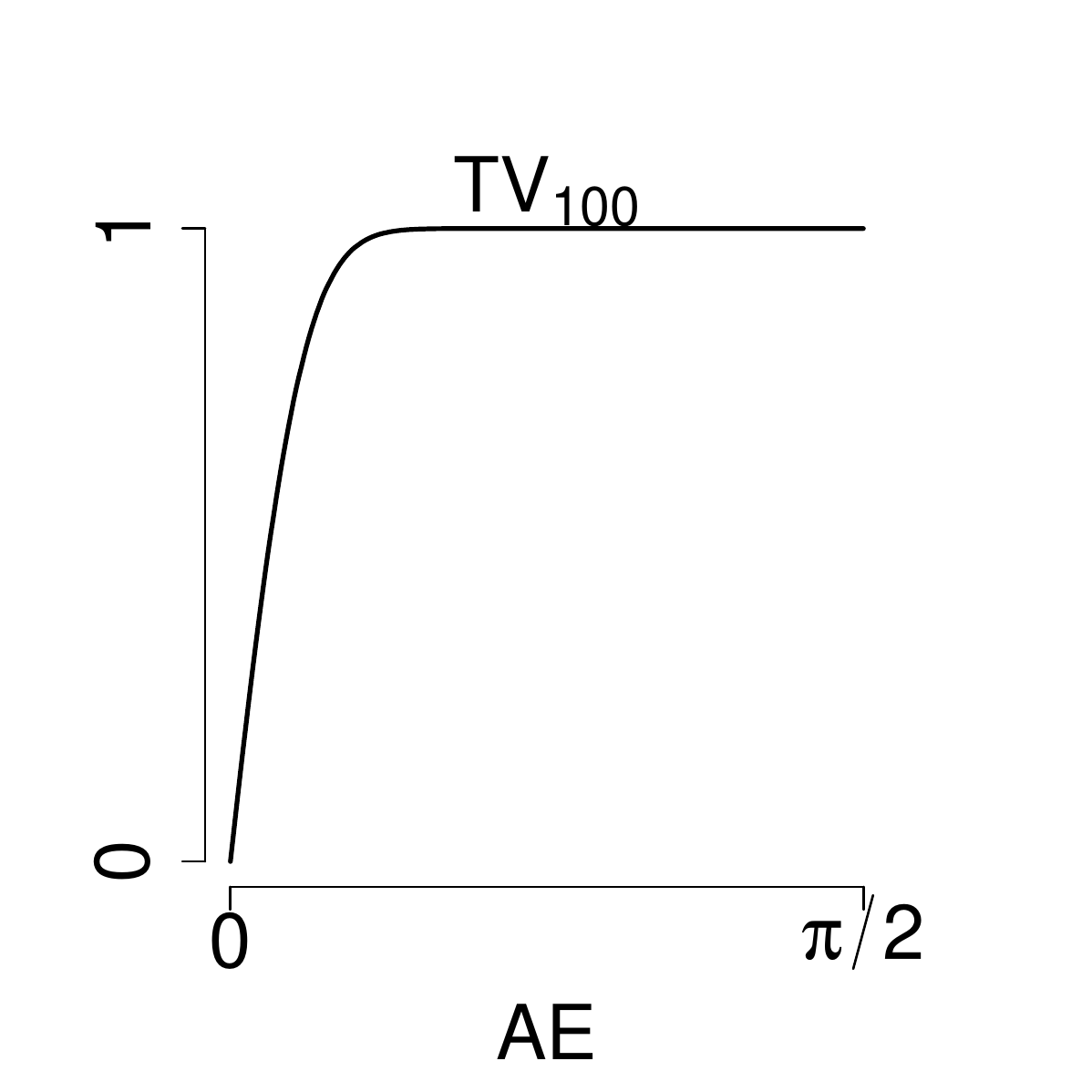}
\caption{Distance between fODFs consisting of a single Dirac delta, $f_1 = \delta_v$, $f_2 = \delta_w$ as a function
of $d_{arc'}(v,w)$.
Left to right: EMD, RMSE, $TV_{gauss,10}$, $TV_{gauss,100}$}
\label{fig:rotation_distance}
\end{figure}

Furthermore,
both the EMD and 2WD equipped with the arc-length distance can be viewed
as an extension of angular error (AE). If we take fODFs consisting of
a single Dirac delta, both the angular error and the EMD distance
between the fODFs is equal to the arc length distance between their
directions: hence in figure \ref{fig:rotation_distance}, we see that
EMD distance is linear with respect to AE; in contrast, RMSE,
$TV_{gauss,1}$ and $TV_{gauss,100}$ are concave with respect to AE.

\subsection{Prediction error}\label{ss:cvrmse}

Unlike other measures of accuracy, the prediction error of a model can
be evaluated without knowing the ground truth, since it uses the
observed data as a noisy version of the ground truth [12].
Furthermore, prediction error can be calculated using a single data
set, via \emph{cross-validation}.

Given data $y_1,\hdots,y_n$ corresponding to measurement directions
$x_1,\hdots,x_n$, one estimates the quantity $\tilde{S}_0$, e.g. by
fitting the NNLS model and setting $\hat{S}_0 = ||\beta||_1$.  The set
of measurement directions $x_1,\hdots,x_n$ is partitioned into $K$
disjoint sets $A_1,\hdots,A_K$ of roughly equal size, and resampled
fODFs $\tilde{f}^1,\hdots,\tilde{f}^K$, where $\hat{f}^i$ are obtained
by estimating the fODF based only on the directions \emph{not} in the
set $A_i$.  Each of the resampled fODFs $\tilde{f}^j$ is used to make
a prediction on the measurements in the \emph{left-out set} $\{y_i: i
\in A_j\}$.  The cross-validated RMSE (CVRMSE) is computed as:
\[
CVRMSE = \sum_{j=1}^K \sum_{i \in A_j} \left(y_i - \hat{S}_0\int_v \exp(-\kappa (x_i' v)^2) d\tilde{f}^j(v)\right)
\]
Alternatively, if two or more replicate measurements $y^1,y^2$ are available,
one can also evaluate the replicated RMSE (RRMSE), defined by $RRMSE =
\sum_{i=1}^n \left(y_i^2 - \hat{S}_0\int_v \exp(-\kappa (x_i' v)^2)
  d\tilde{f}^1(v)\right)$.  The CVRMSE and RRMSE differ only slightly in terms
of mean; the RRMSE has smaller variance.

Supposing that the model is correctly specified, CVRMSE is an nearly
unbiased estimate of the root mean integrated squared error (RMISE)
from the noise-free signal. The RMISE is defined as the $L_2$ distance
between smoothed measures:$d_{RMISE}(f_1,f_2) = \sqrt{\int_w
  (f_{1,ST,\kappa}(v) - f_{2,ST,\kappa}(v))^2 dv}$ where the smoothing
kernel is computed from the Stejskal-Tanner equation [1] with a single
shape parameter $\kappa$: $f_{ST,\kappa}(v) = \int_w
\exp(-\kappa(v'w)^2) df(w)$.

\subsection{Resampled Barycenters}\label{ss:bary}

If one had an accurate Bayesian model of the data, one could obtain an
optimal estimate of the fODF with respect to expected EMD inaccuracy
by obtaining the \emph{Wasserstein barycenter} of the posterior
distribution: 
\begin{equation}\label{bayes_bary}
\hat{f} = \min_{\hat{f}} \int_f d_{EMD}(f,\hat{f}) dp(f|y)
\end{equation}, where $p(f|y)$ is the posterior distribution of the fODF
with respect to the data $y$. The precise form of the posterior
distribution appearing in \eqref{bayes_bary} depends on the particular
prior used. Bayesian approaches for DWI imaging [8] commonly use priors
consisting of mixtures of $K$ dirac deltas, where $K$ also possibly has a prior distribution.
The numerical computation of the Bayesian barycenter
can be achieved by obtaining a large number of posterior samples
$f^1,\hdots,f^N$ from the posterior, then solving
\begin{equation}\label{discrete_bary}
\hat{f} = \min_{\hat{f}} \frac{1}{N}\sum_{i=1}^N d_{EMD}(f^1,\hat{f}) dp(f|y)
\end{equation}
A variety of approaches exist for solving the equation \eqref{discrete_bary},
including linear programming\footnote{
In the case that fODFs $\hat{f}^1,\hdots,\hat{f}^K$ are
mixtures of Dirac deltas, it possible to compute the Wasserstein
barycenter using standard linear program solvers.  Let $v_i^j$ be the
$i$th direction in the fODF $\hat{f}^j$, and $w_i^j$ its corresponding
weight, and let $k_i$ denote the number of directions in $\hat{f}^j$.
Let $u_1,\hdots,u_p$ be a dense sampling on the projective plane.
Then the Wasserstein barycenter is found by the following
optimization problem:
\[
\text{minimize} \sum_{\ell=1}^p \sum_{i=1}^K \sum_{j=1}^{k_i}
x_{\ell i j} d_{arc}(u_\ell, v_i^j)
\]
\[
\text{subject to } \sum_{\ell=1}^p x_{\ell i j} = w_i^j
\text{ , }
w_\ell \geq 0 \text{ , }\sum_{i=1}^K \sum_{j=1}^{k_i} x_{\ell i j} = w_\ell
\]
for $\ell = 1,\hdots,p$, $i = 1,\hdots,K$ and $j = 1,\hdots,k_i$.  The
output of the optimization problem is the values of the variables
$x_{\ell i j}$ for $\ell = 1,\hdots,p$, $i=1,\hdots,K$ and $j=
1,\hdots,k_i$.
The Wasserstein barycenter $\hat{f}_{bary}$ can then be computed as follows.
\begin{enumerate}
\item  Compute $w_1,\hdots,w_p$, by $w_\ell
\sum_{i=1}^K \sum_{j=1}^{k_i} x_{\ell i j}$
\item Let $\hat{f}_{bar} = \sum_{\ell=1}^p w_\ell\delta_{u_\ell}$
\end{enumerate}
}, and a recent approach by Cuturi [14]. However, obtaining the
posterior draws $f^1,\hdots,f^N$ may be extremely expensive.

One can bypass the computational cost of computing the posterior by exploiting the
connection between Bayesian inference and resampling techniques. Efron
[13] demonstrates a close connection between the parametric bootstrap
and Bayesian posteriors for uninformative priors.  The parametric
bootstrap can be immediately applied to our setting: given an
estimated fODF $\hat{f}^0$, and an estimate of the noise
$\hat{\sigma}^2$, generate synthetic bootstrap data $y^1,\hdots,y^K$
by $y^j_i \sim Rician\left(\int_v \exp(-\kappa (v'x_i)^2)
d\hat{f}^0(v), \hat{\sigma}^2\right)$.  Fitting the model to each
synthetic bootstrap replicate $y^1,\hdots,y^K$, obtain bootstrap
estimates of the fODF $\hat{f}^1,\hdots,\hat{f}^K$.  Treating these
bootstrap estimates as a sample from an approximate posterior, compute
$\hat{f} = \min_f \sum_{i=1}^B d_{EMD}(f,\hat{f}^B)$.  An alternate
approach, and one which appears to be more effective in simulations,
is to use $K$-fold partioning rather than parametric bootstrap: that
is, to obtain $\hat{f}^1,\hdots,\hat{f}^K$ using the approach
described in \ref{ss:cvrmse}.

\subsection{K-fold replicate error}

The definition of replicate error requires at least two replicate
measurements of the same voxel, $y^{(1)}$ and $y^{(2)}$: then given a
distance function $d$, the replicate error is defined as $RE =
d(\hat{f}^{(1)},\hat{f}^{(2)})$.  However, one can measure K-fold replicate error (K-RE)
using a single set of measurements by using K-fold partitioning.
Given a single set of measurements $y$, obtain
$\hat{f}^1,\hdots,\hat{f}^K$ according to the $K$-fold partitioning
procedure described in \ref{ss:cvrmse}.  Then define the $K$-fold
replicate error as follows:
\[
\text{K-RE}_d = \frac{K-1}{\sqrt{K}}
\left[\frac{1}{K(K-1)/2}\sum_{1 \leq i < j \leq K}
  d(\hat{f}^i,\hat{f}^j)\right]
\]
The correction factor, $\frac{\sqrt{K}}{K-1}$, is used to reduce the
dependence of the calculated replicate error on the arbitrary choice of
$K$.  Supposing the correction were not employed, the $K$-fold
replicate error would be asymptotically proportional to $\sqrt{(K-1)}/K$,
which is the product of the square root of the relative sample size
$\sqrt{K/K-1}$ and the inverse proportion of directions shared between
different folds, $1/(K-1)$.

\section{Results and Discussion}

\subsection{Comparison of models and accuracy measures}

We compare measures of accuracy applied to simulated estimates of fiber
orientation distribution functions (fODF)
obtained from different models. The measures we consider are angular
error (AE), root mean integrated squared error (RMISE), earth mover's
distance (EMD), total variation (TV) with $\lambda = \{1,10,100\}$ and
symmetric Kullback-Liebler (SKL) with with $\lambda = \{1,10,100\}$\footnote{
The choice of smoothing parameters $\lambda$ for TV and SKL are
somewhat arbitrary; we are not aware of any previous use of smoothed
distances in the DWI literature.}.

\begin{figure}[htbp]
\centering
\includegraphics[scale=.5]{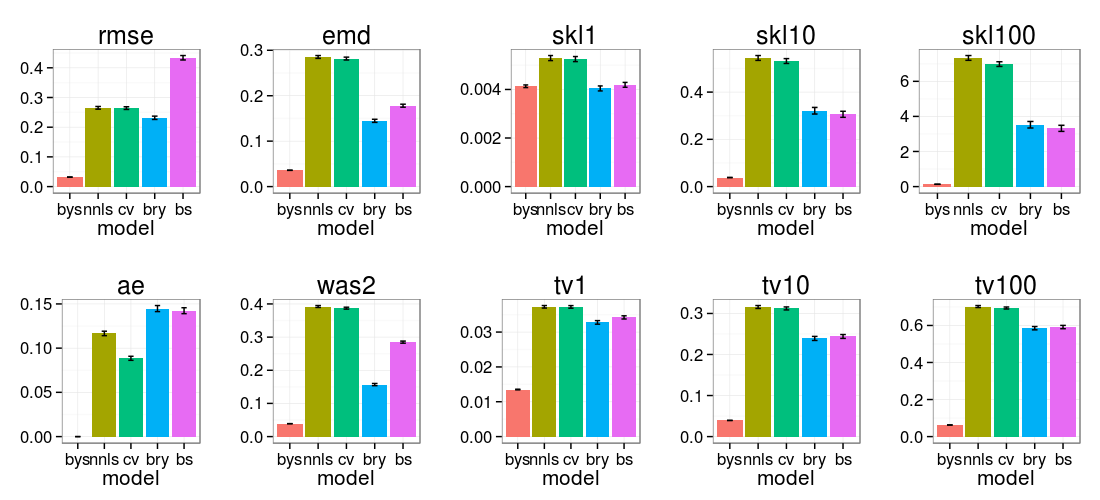}
\caption{Simulated comparison of distance metrics. Models: Bayesian posterior
  mean (bys), Bayesian posterior Wasserstein barycenter (bry),
  best-2-subset (B2S), nonnegative least squares (NNLS), K-fold barycenter
  (cv). For comparison, see also Figure 1.}
\label{fig:table}
\end{figure}

The ground truth fODF consists of two
orthogonal directions with equal weights; the data was generated using
parameters $\kappa=1.5$ and $\sigma^2 = 0.04$.  
These parameters are typical for DWI simulations [2,4,5].
The simulated measurements used measurement directions $x_1,\hdots,x_{150}$ used in
DWI measurements.  We then fit a Bayesian model, best-2-subset and
NNLS.  The Bayesian prior was specified as described in
\ref{ss:models}, and the cross-validated barycenter was computed as
described in \ref{ss:bary} with $K=20$ folds.  Figure
\ref{fig:model_demo} displays sample model fits; Table \ref{fig:table}
provides a table of the measures of accuracy of each model as averaged
over 1000 random trials.

RMISE most strongly favors continuous estimates, such as the Bayes
posterior mean. AE is undefined for continuous estimates and favors
non-sparse estimates, such as NNLS and K-fold barycenter.  On the
opposite side of the spectrum, EMD favors sparse estimates, such as
best-2-subset and the posterior barycenter. TV and SKL do not clearly
favor sparse or non-sparse models.  TV and SKL rank the models
similarly regardless of the smoothing parameter used, but the
smoothing parameter does influence the contrast between different
methods.  In the case of oversmoothing, all models have close to the
minimum inaccuracy, as can be seen in the inaccuracies calculated
using $TV_1$ and $SKL_1$.  In the case of undersmoothing, all models
have close to the maximum inaccuracy, as seen in the inaccuracies
calcuated using $TV_{100}$ and $SKL_{100}$.  In the case of $TV$, we
see that the ratio $\max \text{err}/\min \text{err}$ is equal to 1.3 for $TV_1$, 1.3
for $TV_{10}$, and 1.2 for $TV_{100}$.  In comparison, the ratio $\max
\text{err}/\min \text{err}$ is equal to 1.9 for EMD.

The K-fold barycenter outperforms NNLS in all measures considered
here: a somewhat surprising result, given that the K-fold barycenter
was solely motivated by the goal of minimizing the inaccuracy as
measured by EMD.

\subsection{Correlation of error with replicate error}

\begin{figure}[htbp]
\centering
\includegraphics[scale=.5]{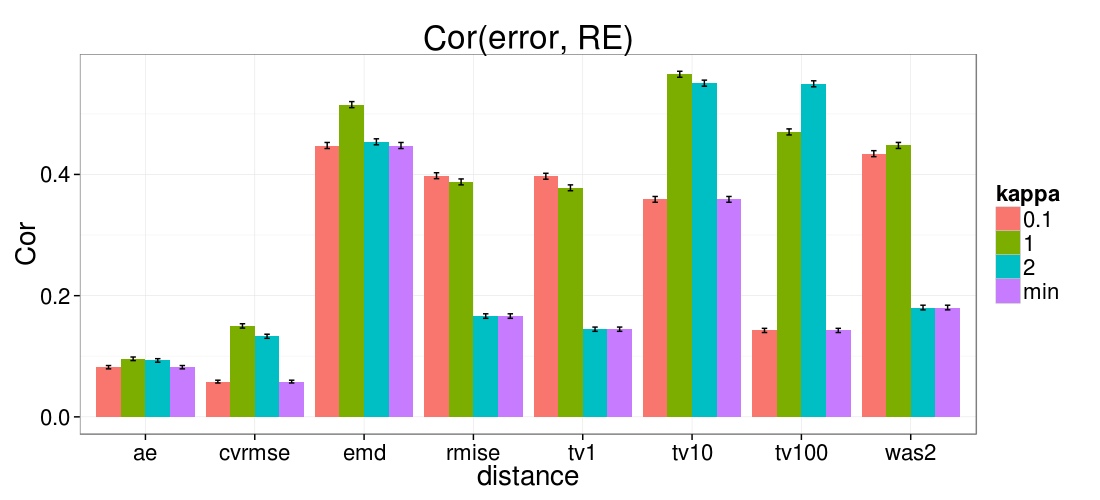}
\caption{Correlation of error and replicate error for data generated from a
  simulation of a two-direction fODF. For each metric, the values at different
  $\kappa$ are presented, and 'min' denotes the minimal correlation among the
  different values of $\kappa$ for that metrix. The column for RRMSE contains values for
  $\text{Corr}(RRMSE,RE_{RMISE})$.  All other columns contain the correlation
  between $\text{err}$ (error) and $RE$ (replicate error) when both quantities are
  evaluated using the given distance/divergence.}
\label{fig:table2}
\end{figure}

In a similar simulated experiment with $k=2$ fiber directions, uniform random
weight $w_1$ and $w_2 = 1-w_1$, $\sigma^2 = 0.04$ and varying $\kappa$, we
compare the correlations between errors $\text{err} = d(f,\hat{f}^1)$ and
replicate errors $RE = d(\hat{f}^1,\hat{f}^2)$.
We find that the correlation between the EMD-based error and EMD-based replicate error,
$\text{Corr}(\text{err}_{EMD},RE_{EMD})$ is above 0.4 for a range of parameter values
$\kappa$ from 0.1 to 2--higher than the minimum correlations for other
distances.  Figure \ref{fig:table2} contains correlations, as computed from
10000 simulations, for several values of $\kappa$.

Given the practical utility of RRMSE, it is surprising to see its low
correlation with the true RMISE regardless of $\kappa$.  Although
RRMSE is a close-to-unbiased estimated of the $\text{err}_{RMISE}$, this may
come at a cost of greater variability.  In contrast, $RE_{RMISE}$ has
a much higher correlation with $\text{err}_{RMSE}$.  At first glance $RE_{RMISE}$ appears to
be a very similar procedure to $RRMSE$, but while $RRMSE$ compares the
signal from replicate 1 with the raw data of replicate 2, $RE_{RMISE}$
compares the signal from replicate 1 with the signal from replicate 2.
The distance measure with the highest correlation between $\text{err}$ and
$RE$ varies depending on $\kappa$. For $\kappa=0.1$, EMD has the
highest correlation, $0.45 \pm 0.01$; for $\kappa = 1$, $TV_1$ has the
highest correlation: $0.55 \pm 0.01$, slightly higher than EMD ($0.52
\pm 0.01$), while for $\kappa = 2$, $TV_{10}$ has the highest
correlation: $0.55 \pm 0.01$.  In both $TV$ and $SKL$ we see that the
choice of smoothing parameter which maximizes the correlation depends
on $\kappa$.  Meanwhile, the 2-Wasserstein distance, which does not
use smoothing, nevertheless has poor correlation between $\text{err}$ and $RE$
at $\kappa=2$, and is consistently dominated by EMD.

To summarize, the correlation of
$\text{Corr}(\text{err}_{EMD},RE_{EMD})$ is consistently comparable to the
highest correlation of any other distance.  Distances with
fixed smoothing kernels suffer from degraded correlation at one of
the extremes of the parameter range, $\kappa = 0.1$ or $\kappa = 2$,
while the 2-Wasserstein distance also suffers from degraded correlation at high $\kappa$ even though it does not employ smoothing;
in contrast, EMD is robust across $\kappa$.

These results can seemingly be explained by the fact that EMD has a
combination of \emph{scale equivariance} and \emph{robustness to
  outliers} as defined in section ~\ref{ss:distances}.  Even though
the two properties were only defined in the Euclidean setting, they
can be extended in a `local' sense to any manifold via the fact that
manifolds resemble Euclidean space in a small neighborhood of any
point.  In the particular application of DWI fiber deconvolution, the
consequence of scale equivariance is that the increased error due to
increased noise level will be reflected both in the error and the
replicate error.  Interestingly, though, we found correlations between
error and replicate error in the simulation even when holding the
noise level fixed.  This can be explained by the fact that even if the noise level
is held fixed, changes in the parameter $\kappa$ or the fiber
configuration can mimic the effect of increased noise.  Thus, the fact
that smoothed TV and smoothed SKL are \emph{not} scale equivariant
explains their inconsistent performance across $\kappa$.  Meanwhile,
the poor performance of 2-Wasserstein distance for $\kappa=2$ can be
explained by the poor robustness of 2-Wasserstein distance to
outliers.  When $\kappa$ is low, relative to the sample size (number of
measurement directions), the NNLS algorithm finds very few `spurious'
directions.  However, when $\kappa$ is high relative to sample size, a
noise spike in a single measurement direction can cause NNLS to weight
an essentially arbitrary direction in a direction orthogonal to the
direction of the noise spike.  This leads to the production of
`outliers' for high $\kappa$, which inflate the variance of the
relative error as measured by 2-Wasserstein distance.  On the other
hand, these directional `artifacts' can be removed by means of
post-processing; hence it would still be interesting to revisit the
application of the 2-Wasserstein distance on post-processed NNLS
estimates.

\subsection{Application to DWI data measured \emph{in vivo} }

DWI data was acquired in a healthy human participant in a 3T MRI instrument, at
the Stanford Center for Neurobiological and Cognitive Imaging. Data was
acquired at 2x2x2 $mm^3$ with a b-value of 2000 s/$mm^2$. The data consists of
two sets of replicate measurements\footnote{The data is available to download
  at: http://purl.stanford.edu/ng782rw8378}.  We identified regions of interest
for analysis in the corpus callosum (CC), a region of the brain known to
contain only one major fascicle, connecting the two cerebral hemispheres, and
in the centrum semiovale (CSO), a part of the brain in which multiple fascicles
cross. We compute K-fold replicate error ($K=10$), and replicate error of the
fODF estimates. We also compute CVRMSE as a direct estimate of accuracy (with
regard to RMISE).  A value of $\kappa = 2.2$ was estimated using
cross-validation on a separate subset of the data.

\begin{figure}[htbp]
\centering
\begin{tabular}{ccc}
K-RE & RE & CVRMSE\\
\includegraphics[scale=0.2]{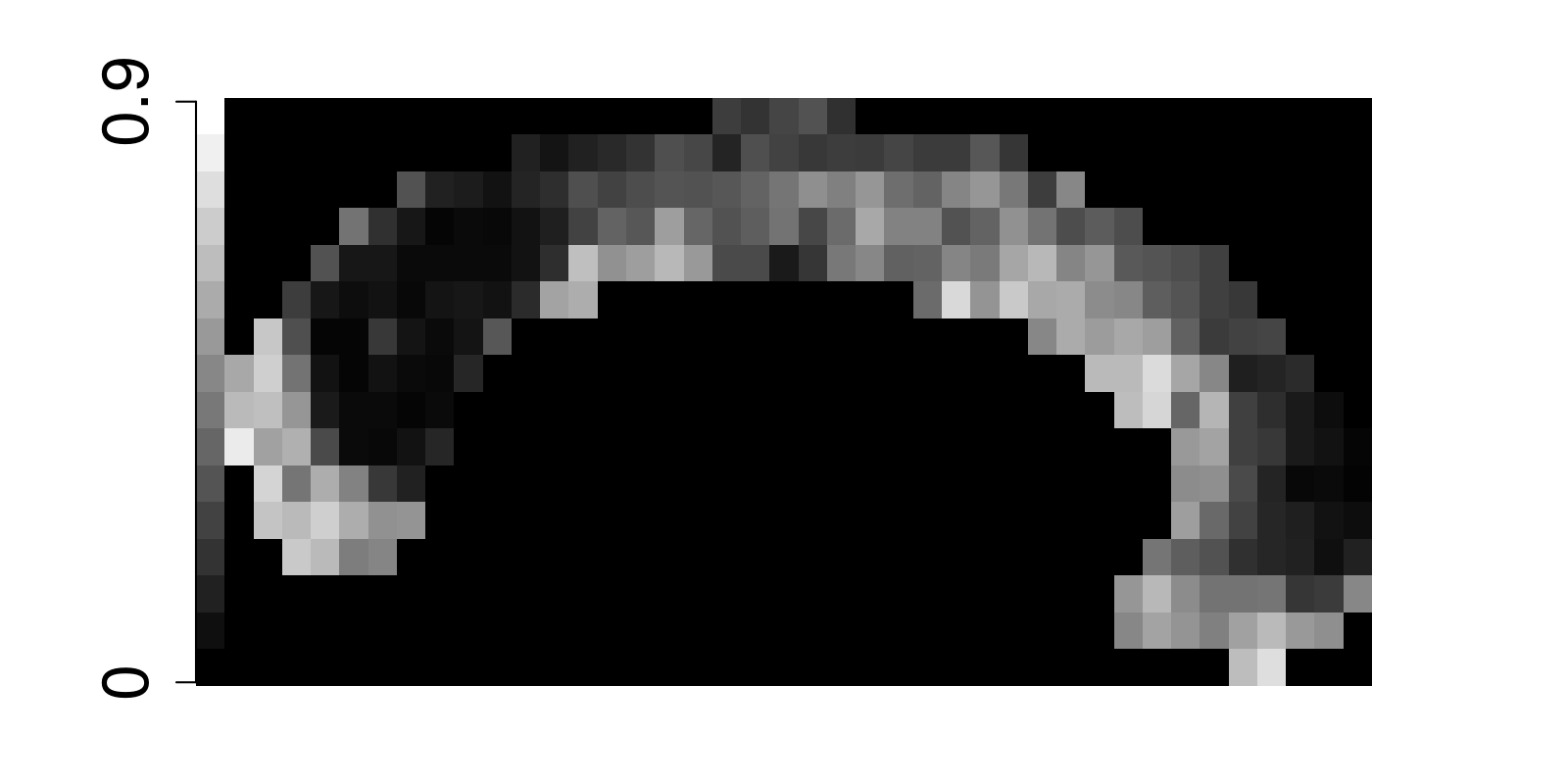} & 
\includegraphics[scale=0.2]{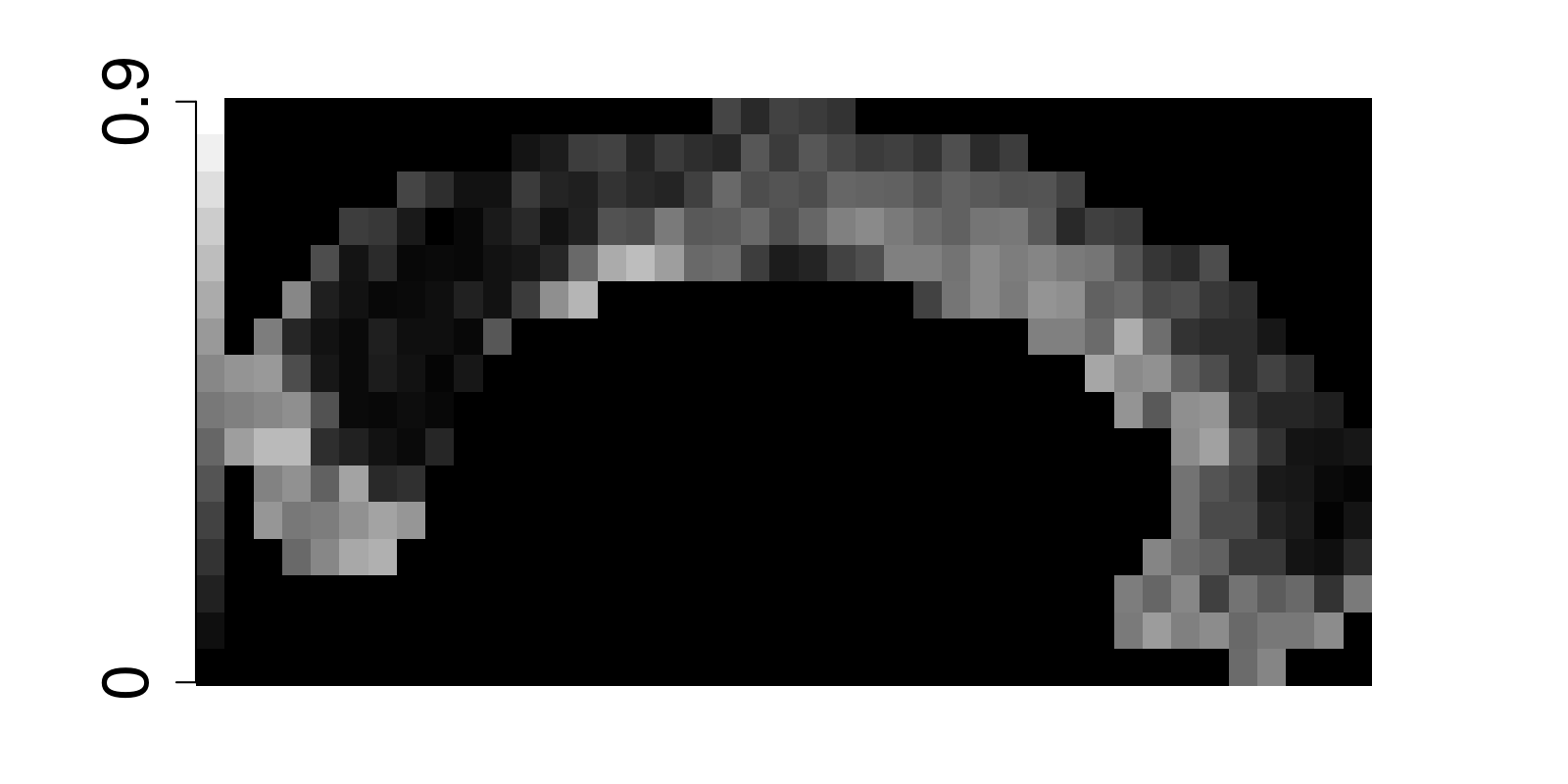} &
\includegraphics[scale=0.2]{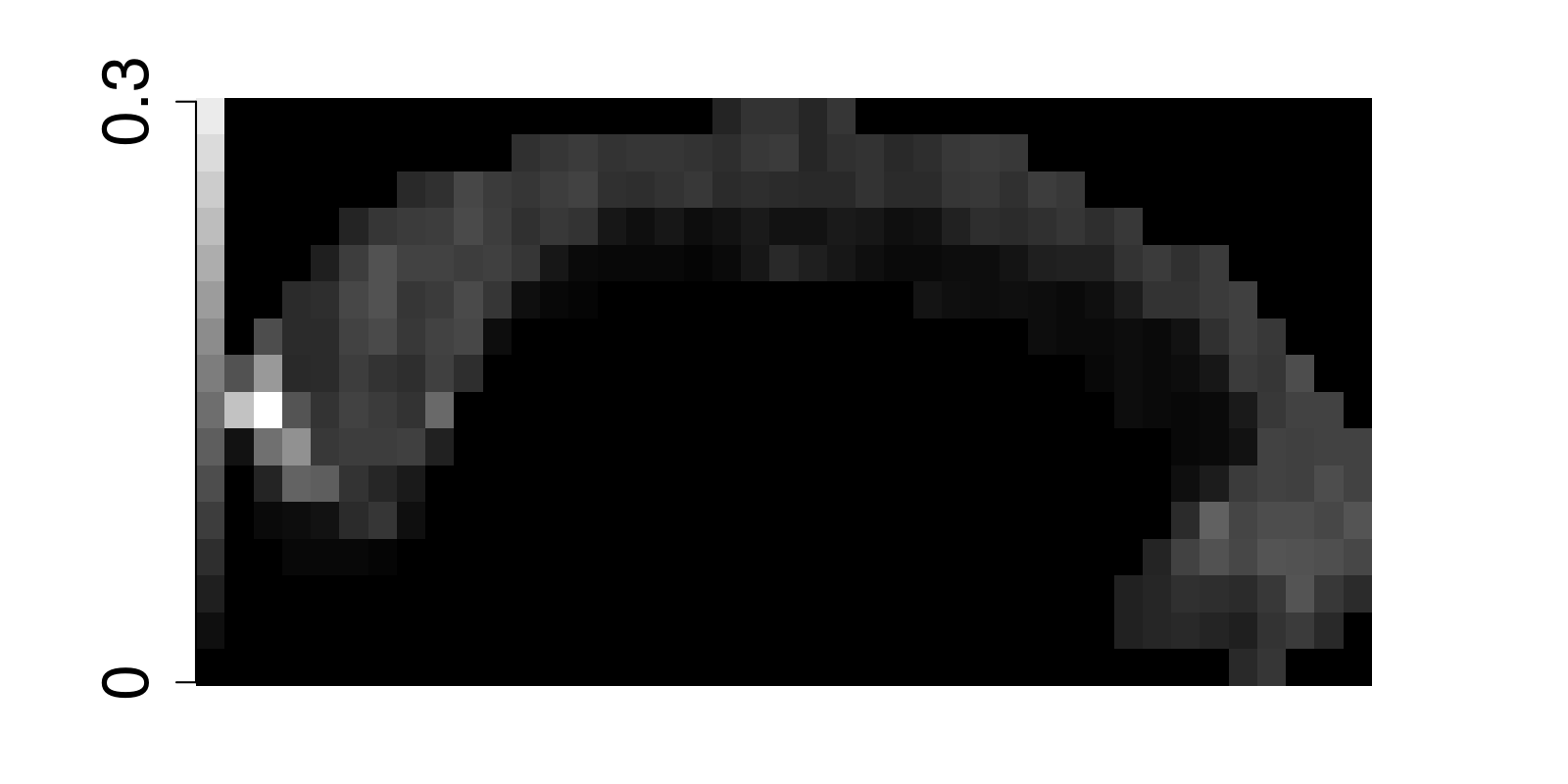}
\\
\includegraphics[scale=0.2]{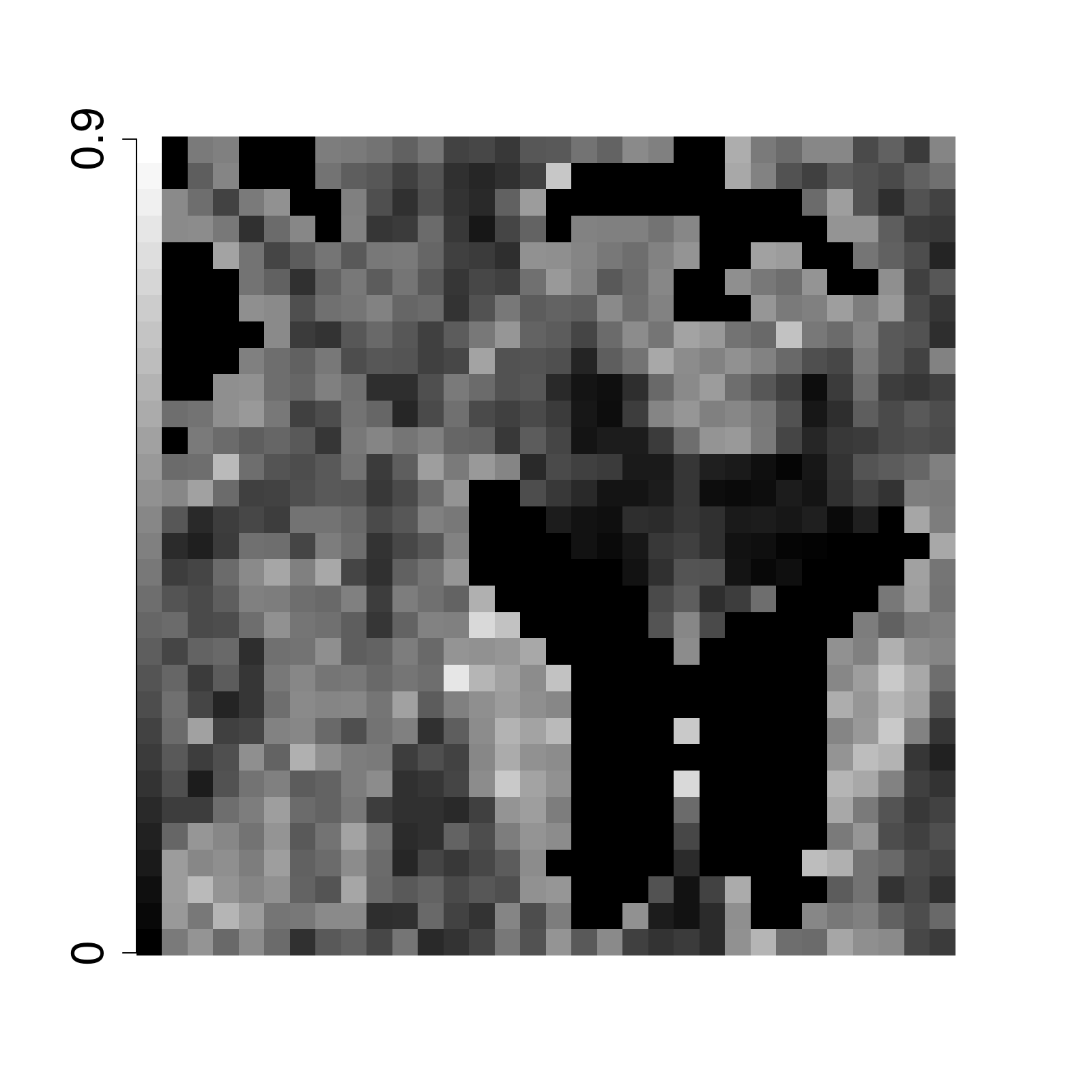} & 
\includegraphics[scale=0.2]{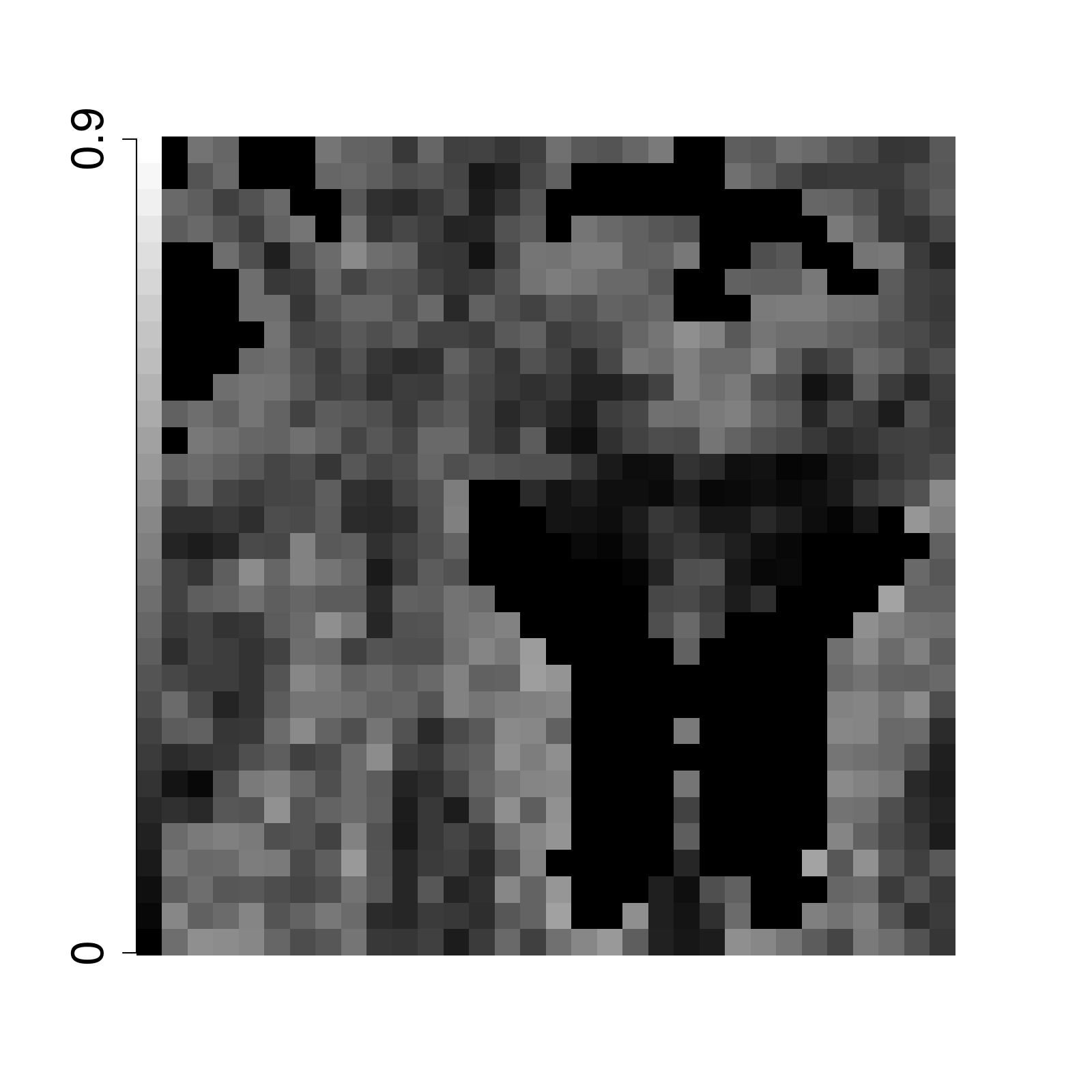} &
\includegraphics[scale=0.2]{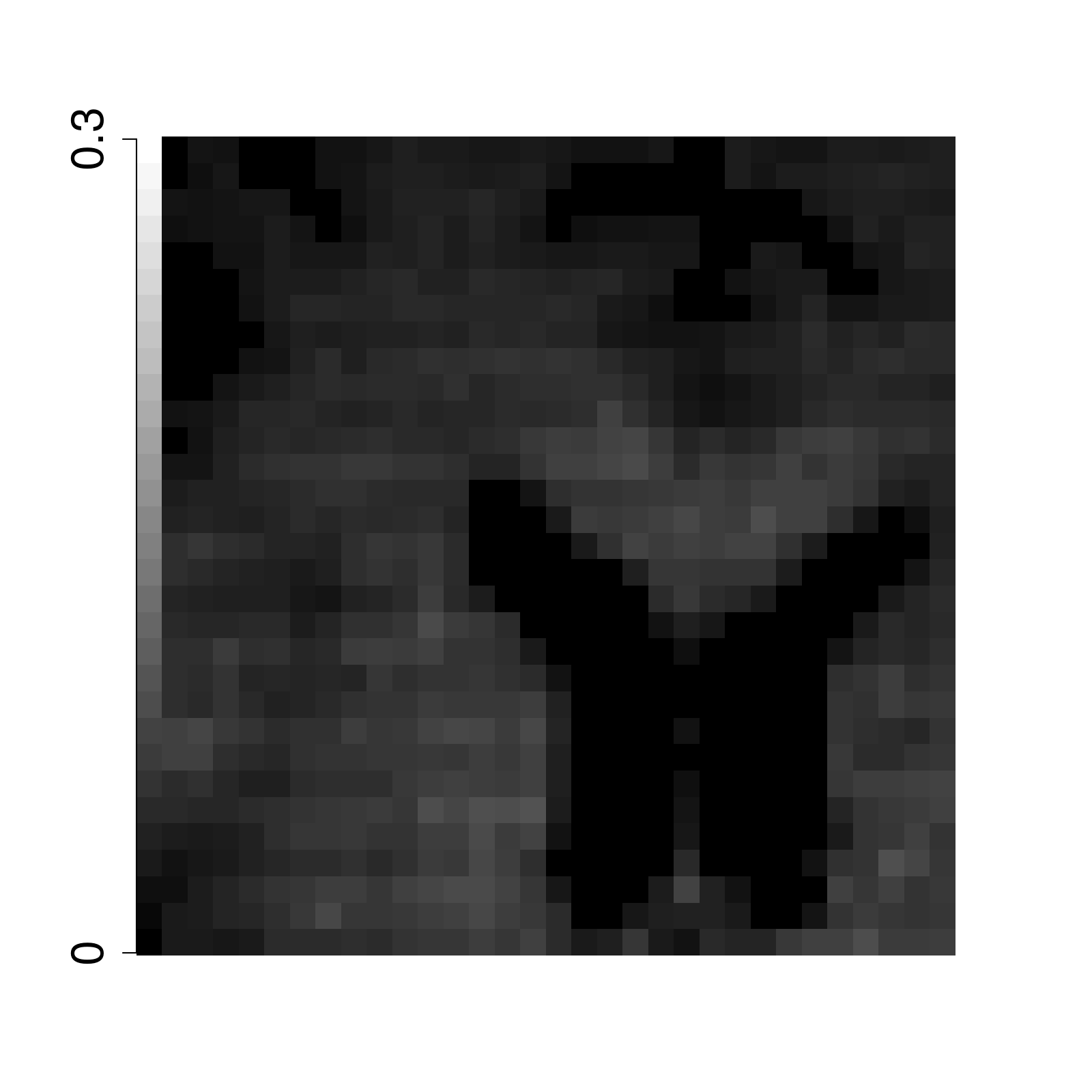}
\end{tabular}
\caption{Empirical measures of replicate error (EMD) and prediction
  error (CVRMSE) by brain region. Top row: Sagittal view of the corpus
  callosum (CC), a part of the brain that contains fibers connecting
  the two cerebral hemsipheres. Bottom row: Axial view of the centrum
  semiovale (CSO), a part of the brain that contains multiple crossing
  fiber populations}
\label{fig:cvemd}
\end{figure}

CVRMSE appears to vary more smoothly across both regions of interest
in the white matter. On the other hand, measures of replicate error
(both RE and K-RE) show more coherent spatial variation.  Both CVRMSE
and replicate error are sensitive both to the configuration of the
fibers in the measurement voxel and to the noise in the measurement,
but their sensitivities to these factors differ. While CVRMSE is
primarily sensitive to noise, EMD-based replicate error is more
sensitive to the configuration of the underlying tissue (i.e. single
fiber population, or more populations of fibers).  The spatial
variations in EMD across the corpus callosum ROI represent, therefore,
variations in the degree to which different parts of the measurement
contain partial volumes of other neighboring parts of the
tissue. These other parts may contain either cerebrospinal fluid (the fluid the
surrounds and pervades the brain), or
fibers oriented in other directions than the corpus callosum
fibers. The measurement noise, on the other hand, is dominated by
physiological factors, and instrumental factors that vary very little
across space. Hence, the relative smoothness of the variation of
CVRMSE across these regions.

\section{Conclusions}

In this paper we address the question of selecting an error metric for
fODF estimation. Through simulations, we illustrate the differences
between EMD and alternative metrics, such as smoothed total variation
and RMISE.  EMD favors sparse estimates of the fODF, and is an
intuitive extension of angular error, which is commonly used to
characterize accuracy in the DWI literature. These properties favor
the use of EMD in theoretical work and simulations. In practice, one
might only be able to measure replicate error, or K-fold replicate
error. Use of the EMD in practical applications, on empirical data, is
supported by the consistent correlation of approximately 0.4 between
replicate error and error as measured by EMD across a wide range of
experimental conditions and biological factors (embodied in the model
parameterization by $\kappa$). Other error metrics, such as smoothed
total variation distance, may have higher correlation between
replicate error and error, but this depends on the smoothing
parameter $\lambda$.  EMD has a unique combination of scale
equivariance and robustness to outliers, which further supports the
use of EMD-based replicate error as a proxy for EMD-based error. The
use of EMD as an error metric motivates the use of Wasserstein
barycenters as estimates of the fODF: while the K-fold barycenter is
motivated as an approximation to the Bayesian posterior barycenter, we
find in simulations that the K-fold barycenter outperforms NNLS in all
measures of accuracy considered, hence meriting more detailed
investigation of its properties.

\subsubsection*{Acknowledgments}
The authors thank Trevor Hastie, Brian Wandell, Eero Simoncelli,
Justin Solomon, Leo Guibas and Shuo Xie for useful discussions, and
the anonymous referees for their helpful suggestions. CZ was supported
through an NIH grant 1T32GM096982 to Robert Tibshirani and Chiara
Sabatti, AR was supported through NIH fellowship F32-EY022294. FP was
supported through NSF grant BCS1228397 to Brian Wandell

\subsubsection*{References}

[1] Le Bihan D, Mangin JF, Poupon C, Clark CA, Pappata
S, Molko N, Chabriat H. (2001). Diffusion tensor imaging:
concepts and applications. \emph{Journal of magnetic resonance imaging},
13(4), 534-546.

[2] Tournier J-D, Calamante F, Connelly A (2007). Robust determination of the
fibre orientation distribution in diffusion MRI: non-negativity constrained
super-resolved spherical deconvolution. {\it Neuroimage} 35:1459–72

[3] Tournier, J.-D., Yeh, C.-H., Calamante, F., Cho, K.-H., Connelly, A., and
Lin, C.-P. (2008). Resolving crossing fibres using constrained spherical
deconvolution: validation using diffusion-weighted imaging phantom
data. NeuroImage, 42: 617–25.

[4] Basser PJ. Quantifying errors in fiber-tract direction and diffusion tensor
field maps resulting from MR noise. Proc. Int. Soc. Magn. Reson. Med. 1997

[5] Aganj I, Lenglet C, Jahanshad N, Yacoub E, Harel N, Thompson PM,
Sapiro G. (2011). A Hough transform global probabilistic approach to
multiple-subject diffusion MRI tractography. \emph{Medical image
  analysis}, 15(4), 414-425.

[6] Frank L. Anisotropy in high angular resolution diffusion-weighted
MRI.  \emph{Magnetic Resonance in Medicine} Volume 45, Issue 6, pages
935–939, June 2001

[7] Dell’Acqua F, Rizzo G, Scifo P, Clarke RA, Scotti G, Fazio F (2007). A
model-based deconvolution approach to solve fiber crossing in
diffusion-weighted MR imaging. {\it IEEE Trans Biomed Eng} 54:462–72

[8] Behrens TEJ, Berg HJ, Jbabdi S, Rushworth MFS, and Woolrich MW (2007).  Probabilistic
diffusion tractography with multiple fiber orientations: What can we
gain?  {\it NeuroImage} (34):144-45.

[9] Gudbjartsson, H., and Patz, S. (1995). The Rician distribution of noisy MR
data. {\it Magn Reson Med}. 34: 910–914.

[10] Parzen E. On the estimation of a probability density fuction and mode.
\emph{The Annals of Mathematical Statistics}. 33(3): 1065-1076, 1962.

[11] Rubner, Y., Tomasi, C.  Guibas, L.J.  (2000).  The earth mover's distance as a
metric for image retrieval.  {\it Intl J. Computer Vision}, 40(2), 99-121.

[12] Rokem A, Yeatman J, Pestilli F, Kay K, Mezer A, van der Welt S,
Wandell B.  (2013). Evaluating the accuracy of models of diffusion MRI
in white matter.  Submitted.

[13] Efron B.  Bayesian inference and the parametric bootstrap.
\emph{The Annals of Applied Statistics} 6 (2012), no. 4,
1971--1997. 

[14] Cuturi M, Doucet A.  Fast computation of Wasserstein
barycenters. \emph{JMLR W\&CP} 32 (1) : 685–693, 2014

\end{document}